\documentclass[10pt,twocolumn,letterpaper]{article}

\usepackage{cvpr}
\usepackage{times}
\usepackage{epsfig}
\usepackage{graphicx}
\usepackage{amsmath}
\usepackage{amssymb}

\usepackage{epstopdf}
\usepackage{epsfig,mdwlist}
\usepackage{amsmath}
\usepackage{amsfonts}
\usepackage{arydshln}
\usepackage{subfigure}
\usepackage{amsmath}
\usepackage{breqn}
\usepackage{multirow}
\usepackage{threeparttable}

\usepackage{appendix}

\usepackage{colortbl}
\definecolor{mygray}{gray}{0.9}

\usepackage[pagebackref=true,breaklinks=true,letterpaper=true,colorlinks,bookmarks=false]{hyperref}

\cvprfinalcopy % *** Uncomment this line for the final submission

\ifcvprfinal\fi
\begin{document}

%%%%%%%%% TITLE
\title{Exploiting temporal and depth information for multi-frame face anti-spoofing}

\author{Zezheng Wang$^1$, Chenxu Zhao$^{1}$\footnotemark[1]\ , Yunxiao Qin$^1$, Qiusheng Zhou$^1$, \\ Guojun Qi$^2$, Jun Wan$^3$, Zhen Lei$^3$ \\
$^1$JD Finance \\
$^2$Huawei Cloud\\
$^3$Chinese Academy of Sciences \\
\{wangzezheng1, zhaochenxu1, qinyunxiao, zhouqiusheng\}jd.com, guojunq@gmail.com, \\
\{jun.wan, zlei\}@nlpr.ia.ac.cn
}

\maketitle

\renewcommand{\thefootnote}{\fnsymbol{footnote}}
\footnotetext[1]{ denotes corresponding author.}

%%%%%%%%% ABSTRACT
\begin{abstract}
   Face anti-spoofing is significant to the security of face recognition systems. Previous works on depth supervised learning have proved the effectiveness for face anti-spoofing. Nevertheless, they only considered the depth as an auxiliary supervision in the single frame. Different from these methods, we develop a new method to estimate depth information from multiple RGB frames and propose a depth-supervised architecture which can efficiently encodes spatiotemporal information for presentation attack detection. It includes two novel modules: optical flow guided feature block (OFFB) and convolution gated recurrent units (ConvGRU) module, which are designed to extract short-term and long-term motion to discriminate living and spoofing faces. Extensive experiments demonstrate that the proposed approach achieves state-of-the-art results on four benchmark datasets, namely OULU-NPU, SiW, CASIA-MFSD, and Replay-Attack.
\end{abstract}

%%%%%%%%% BODY TEXT
\section{Introduction}
\label{sec:intro}
Face recognition systems have become indispensable in many interactive AI systems for its convenience. However, most of existing face recognition systems are so vulnerable to the face spoofing that attackers can easily deceive face recognition systems by using presentation attacks (PA), e.g., printing a face on paper (print attack), replaying a face on a digital device (replay attack), and bringing a 3D-mask (3D-mask attack). Such PAs can tamper with face recognition systems, placing direct risks to money payment and privacy verification that have high stakes with the public interests. Therefore, face anti-spoofing plays a critically important role in the security of face recognition.

To defend against face spoofing attacks, a great number of face anti-spoofing methods~\cite{Boulkenafet2017Face,Gan20173D,Lucena2017Transfer, Nagpal2018A} have been proposed to discriminate the living and fake face. Previous approaches fall into two categories. The first is the traditional methods, which commonly train shallow classifiers with hand-crafted features, i.e., LBP~\cite{Pereira2012LBP}, SIFT~\cite{Patel2016Secure}, or SURF~\cite{Boulkenafet2017Face_SURF}. With only texture characteristic considered, these methods are often prone to the attacks, such as replaying mediums and 2D/3D mask.
\begin{figure}[t]
  \includegraphics[width=0.48\textwidth]{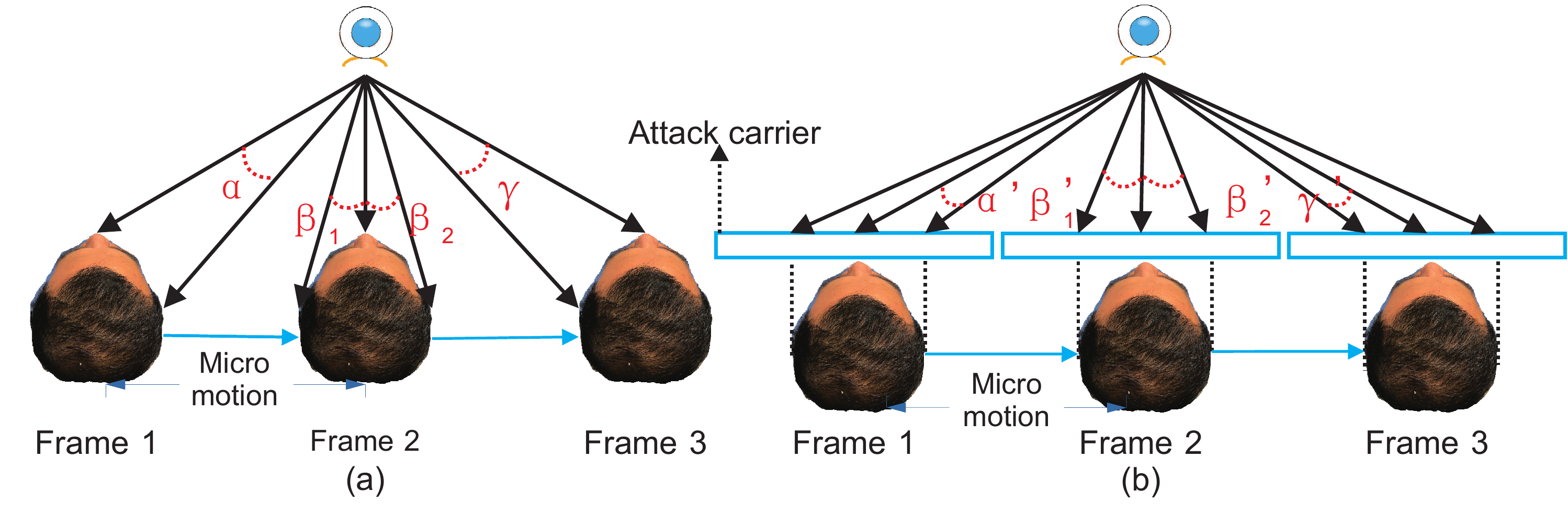}
  \caption{%The figure shows different viewpoints in live and spoof(print attack here) scenes.
  To observe the difference between live and spoof(print attack here) scenes, we represent the micro motion exaggeratedly. The angle of camera viewpoint can reflect the facial motion relative depth among different keypoints. In the living scene (a), the angle $\alpha$ between nose and right ear is getting smaller with the face moving right ($\alpha > {\beta}_2$), while the angle ${\beta}_2$ between left ear and nose is getting larger (${\beta}_1 < {\gamma}$). However, in the spoofing scene (b), $\alpha'< {\beta}'_2$, and ${\beta}'_1 > {\gamma}'$. %Even the camera can see the left ear at frame 1 and right ear at frame 3, which are in the blind zone of living scene.
  }
  \label{fig:intro}
\vspace{-10pt}
\end{figure}
In contrast to texture-based methods, CNN-based methods provide an alternative way to learn discriminative anti-spoofing representations in an end-to-end fashion. Some CNN-based methods~\cite{Lucena2017Transfer, Nagpal2018A, Gan20173D, Xu2016Learning, Yang2014Learn, Shao2017Deep} only treat face anti-spoofing as a pure binary classification - spoofing as 0 and living as 1, and train the neural network supervised by simple softmax loss. However, these methods fail to explore the nature of spoof patterns~\cite{Liu2018Learning}, which consist of skin detail loss, color distortion, moir$\rm\acute{e}$ pattern, motion pattern, shape deformation and spoofing artifacts.

Depth supervised face anti-spoofing somehow relieves these issues.
%To compare the real scenes of living scene with spoofing scene
Intuitively, images of living faces contain face-like depth information, whereas spoofing face images in print and replay carriers only have planar depth information. Thus, Atoum \etal.~\cite{Atoum2018Face} and Liu \etal. ~\cite{Liu2018Learning} proposes single-frame depth supervised CNN architecture in their architecture, and improves the performance of presentation attack detection (PAD). %Both of them utilized facial depth as supervision and designed an FCN structure to regress single-frame facial depth.
However, they only leverage facial depth as an auxiliary strong supervision to detect the static spoofing cues under pixel-wise supervision relatively carefully, neglecting to exploit the \emph{virtual} depth difference between living and spoofing scenes. %In their work, facial depth is more likely a tool to force the network to detect the static spoofing cues under pixel-wise supervision more carefully.
To compensate with the lack of temporal information, Liu \etal.~\cite{Liu2018Learning} takes rPPG signals as extra supervisory cues, and does not use any depth information in temporal domain. However, depth information is more obvious in sequential frames. Indeed, temporal motion is very useful to extract depth information and thus play a great potential role for face anti-spoofing detection.

In this paper, we will explore the temporal depth by combining temporal motion and facial depth, and demonstrate that face anti-spoofing can be significantly benefit from temporal facial depth. No matter whether cameras or subjects move, sequential frames can be represented in a 3D space using the relative motion of objects. Inspired by the work~\cite{Shahraray1988Robust} revealing the relationship between relative facial depth and the motion, we believe patterns of motion and depth are distinct between living and spoofing scenes. And the abnormal facial depth will reflect on the unique facial motion pattern in the temporal domain. Fig.~\ref{fig:intro} illustrates apparently different temporal representations between living and spoofing scenes. %We find that it's powerful to use temporal information under depth supervision to detect the presentation attacks.
To this end, we present a novel depth supervised neural network architecture with optical flow guided feature block (OFFB) and convolution gated recurrent units (ConvGRU) module to incorporate the temporal information from both adjacent and long-sequence frames, respectively. %Because the motion in face recognition may be micro.
By combining both short-term and long-term motion extractors, the temporal motion can be used effectively to discriminate living and spoofing faces under depth supervision. And to estimate facial depth in face anti-spoofing tasks, we propose a novel contrastive depth loss to learn topography of facial points.

We summarize the main contributions below:
\vspace{-6pt}
\begin{itemize}
\setlength{\itemsep}{0pt}
\setlength{\parsep}{0pt}
\setlength{\parskip}{0pt}
\item We analyze the temporal depth in face anti-spoofing and seek the usefulness of motion and depth in  face anti-spoofing.
%第一个创新点，应该是提出来一种**方法，利用多帧运动信息，求解depth的方法
\item We propose a novel depth supervised architecture with OFF block (OFFB) and ConvGRU module to uncover facial depths and their unique motion patterns from temporal information of monocular frame sequences. %significant temporal information and detect the exceptional facial depth.
%第二个创新点：说一下OFF block 和 congru在这个框架里面的作用
\item We design a contrastive depth loss to learn the topography of facial points for depth supervised face anti-spoofing.
\item We demonstrate the superior performances to the state-of-the-art methods on widely used face anti-spoofing benchmarks.%achieve the state-of-the-art performances on standard face anti-spoofing benchmarks.
\end{itemize}

\section{Related Work}
\label{sec:related}
We review related face anti-spoofing works in three categories: binary supervised methods, depth supervised methods, and temporal-based methods.

\noindent \textbf{Binary supervised Methods} Since face anti-spoofing is essentially a binary classification problem, most of previous anti-spoofing methods purely train a classifier under binary supervision, e.g., spoofing face as 0 and living face as 1. Binary classifiers contain traditional classifiers and neural networks. Prior works usually rely on hand-crafted features, such as LBP~\cite{Pereira2012LBP,Pereira2013Can,Maatta2011Face}, SIFT~\cite{Patel2016Secure}, SURF~\cite{Boulkenafet2017Face_SURF}, HoG~\cite{Komulainen2014Context,Yang2013Face}, DoG~\cite{Peixoto2011Face,Tan2010Face} with traditional classifiers, such as SVM and Random Forest. Since these manually-engineered features are often sensitive to varying condition, such as camera devices, lighting conditions and presentation attack instruments (PAIs), traditional methods often perform poorly in generalization.

%As the development of hardware and the increase of the amount of data, CNN has proven to be a successful method in a number of computer vision problems.
Since then, CNN has achieved great breakthrough with the help of hardware development and data abundance.
Recently, CNN is also widely used in face anti-spoofing tasks~\cite{Lucena2017Transfer,Nagpal2018A,Gan20173D,Feng2016Integration,Li2017An,Patel2016Cross,Yang2014Learn}. However, most of the deep learning methods simply consider face anti-spoofing as a binary classification problem with softmax loss. Both ~\cite{Li2017An} and ~\cite{Patel2016Cross} fine-tune a pre-trained VGG-face model and take it as a feature extractor for the subsequent classification. Nagpal \etal.~\cite{Nagpal2018A} comprehensively study the influence of different network architectures and hyperparameters on face anti-spoofing. Feng~\cite{Feng2016Integration} and Li~\cite{Li2017An} feed different kinds of face images into the CNN network to learn discriminative features on living faces and spoofing faces.

\noindent \textbf{Depth supervised Methods} Compared with binary supervised face anti-spoofing methods, depth supervised methods have a lot of advantages. Atoum~\cite{Atoum2018Face} utilizes depth map of face as supervisory signals for the first time. They propose a two-stream CNN-based approach for face anti-spoofing, by extracting the local features and holistic depth maps from the face images. In other words, they combine depth-based CNN and patch-based CNN from single frame to obtain discriminative representation to distinguish live vs. spoof. It shows that depth estimation is beneficial for modeling face anti-spoofing to obtain promising results especially on higher-resolution images.

Liu~\cite{Liu2018Learning} proposes a face anti-spoofing method from a combination of spatial perspective (depth) and temporal perspective (rPPG). They regard facial depth as an auxiliary supervision, along with rPPG signals. For temporal information, they use a simple RNN to learn the corresponding rPPG signals. However, due to the simplicity of sequence-processing, they have a non-rigid registration layer to remove the influence of facial poses and expressions, ignoring that unnatural changes of facial poses or expressions are significant spoofing cues.

\noindent \textbf{Temporal-based Methods} Temporal information plays a vital role in face anti-spoofing tasks. ~\cite{Pan2007Eyeblink,Patel2016Cross,Shao2017Deep} focus on the movement of key parts of the face. For example, ~\cite{Pan2007Eyeblink,Patel2016Cross} make spoofing predictions based on eye-blinking. These methods are vulnerable to replay attack since excessively relying on one aspect. Gan~\cite{Gan20173D} proposes a 3D convolution network to distinguish the live vs. spoof. 3D convolution network is a stacked structure to learn the temporal features in a supervised pattern, but depends on significant amount of data and performs poorly on small database. Xu~\cite{Xu2016Learning} proposes an architecture combining LSTM units with CNN for binary classification. Feng~\cite{Feng2016Integration} presents a work that takes optical flow magnitude map and Shearlet feature as inputs to CNN. Their work shows that optical flow map presents obvious difference between living faces and different kinds of spoofing faces. All prior temporal-based methods are incapable of catching valid temporal information with a well-designed structure. %In order to conquer this problem, we propose a neural network which combines the short-term motion module and long-term motion module to seek excellent facial motion to estimate facial depth and to detect PAs.

\section{The Proposed Approach}
%In our proposed approach, we make rational use of temporal information to recover facial depth, which is used as a supervision signal to catch spoof patterns for face anti-spoofing.
In this section, firstly, we will introduce the temporal depth in face anti-spoofing and show the useful patterns of motion and depth in face anti-spoofing. Then we will present the proposed temporal PAD method under depth supervision. As shown in Fig.~\ref{fig:pipeline}, the proposed model mainly consists of two modules. One is the single-frame part, focusing on seeking spoofing cues with the static depth supervision. The other is the multi-frame part, consisting of optical flow guided feature block (OFFB) as the short-term motion module and convolutional gated recurrent units (ConvGRU) modeling long-term motion patterns. The model can explore spoofing cues in both spatial and temporal domains under the depth supervision, successfully.
\subsection{Temporal Depth in Face Anti-spoofing}
\label{subsec:theory}
\begin{figure}[t]
  \includegraphics[width=0.48\textwidth]{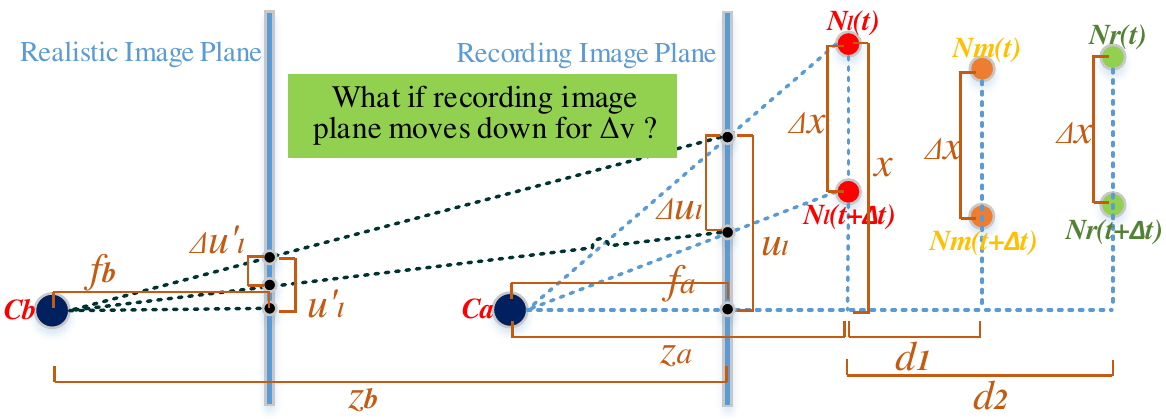}
  \caption{The schematic diagram of motion and depth variation in the presentation attack scene.
  }
  \label{fig:theory1}
\end{figure}
As shown in Fig.~\ref{fig:theory1}, there are two image spaces. The right image space is recording image space, where attackers use to record others' faces. The left image space is realistic image space, where real face recognition systems are. In the figure, $C_a, C_b$ are the focal points of recording image space and realistic image space, respectively, $f_a, f_b$ are corresponding focal distances, and $z_b$ is the distance between $C_b$ and the recording image plane (i.e., the screen of replay attack or the paper of print attack), while $z_a$ is the distance between the focal point and the nearest facial point. $N_l(t), N_m(t), N_r(t)$ are three points having different distances to the camera at time $t$, and $N_l(t+\Delta t), N_m(t+\Delta t), N_r(t+\Delta t)$ are the corresponding points when they move down vertically by $\Delta x$ at time $t+\Delta t$. $x$ is the coordinate in the vertical direction, and $u_l$ is the mapping coordinate in the recording image plane. $\Delta u_l$ represents the optical flow on the image plane when the point moves down. $u'_l, \Delta u'_l$ are values mapped on the realistic image plane by $u_l, \Delta u_l$, respectively. We have the following two equations revealing the relation between these variables:
\begin{equation}
\begin{split}
\frac{\Delta u'_l}{\Delta u'_m} = \frac{d'_1}{z_b} + 1,\\
\frac{\Delta u'_l}{\Delta u'_r} = \frac{d'_2}{z_b} + 1,
\end{split}
\label{eq:theory1:1}
\end{equation}
where just like $\Delta u'_l$, $\Delta u'_m$ and $\Delta u'_r$ are optical flow of $N_m(t)$ and $N_r(t)$ mapping on the realistic image plane, respectively, which are not shown in the figure for simplicity. And $d'_1$ and $d'_2$ are estimated depth difference $d_1$ and $d_2$, that can be calculated by the optical flow. The other is
\begin{equation}
\begin{split}
\frac{d'_1}{d'_2} = \frac{d_1}{d_2} \cdot \frac{f_a \Delta x + (z_a + d_2) \Delta v} {f_a \Delta x + (z_a + d_1) \Delta v},
\end{split}
\label{eq:theory1:2}
\end{equation}
where $\Delta v$ is the distance of the recording image plane moving down vertically, caused by the shake of attack carrier. Here, we only consider vertical direction of movement.
From these relations, we have some important conclusions:
\begin{itemize}
\setlength{\itemsep}{0pt}
\setlength{\parsep}{0pt}
\setlength{\parskip}{0pt}
\item In real scene, we can obtain the correct estimate of relative depth: $d'_1/d'_2 = d_1/d_2$.
\item In print attack scene, $\Delta u'_l = \Delta u'_m,\ \Delta u'_l = \Delta u'_r$. Then we can use Eq.~\ref{eq:theory1:1} to show $d'_1 = 0,\ d'_2 = 0$, which suggests the face is a plane.
\item In replay attack scene, if $\Delta v = 0$, then $d'_1/d'_2 = d_1/d_2$. In this case, there is no abnormity of relative depth. In the meantime, if there are no static spoofing cues , we call this case \textbf{Perfect Spoofing Scene(PSS)}. However, making up \textbf{PSS} attacks is costly and approximately impossible in practice. Thus we usually assume $\Delta v \neq 0$, where $d'_1/d'_2 \neq d_1/d_2$. The estimated relative depth in attacking scenes is different from that in real scenes. And $\Delta v$ as well as $d'_1/d'_2$ usually varies in the long-term sequence. Such variations can also be taken as a valid spoofing cue.
\item If $d_2$ denotes the largest depth difference among facial points, then $d_1/d_2 \in [0, 1]$, showing that constraining depth label of living face to $[0, 1]$ is valid. For spoofing scenes, the abnormal relative depth is too complex to be computed directly. Therefore, we merely set depth label of spoofing face to all 0 to distinguish it from living label, making the model learn the abnormity under depth supervision itself.
\end{itemize}
More details about the derivation are included in the supplementary material.

\subsection{Single-frame Part}
Single-frame part is important to learn static spoofing information. In this work, we train a simple CNN to regress the depth map instead of traditional classification with softmax loss. In the following, we introduce the proposed depth supervised single-frame architecture in two aspects: depth generation and network structure.
\subsubsection{Depth Generation}
\begin{figure}[t]
  \includegraphics[width=0.48\textwidth]{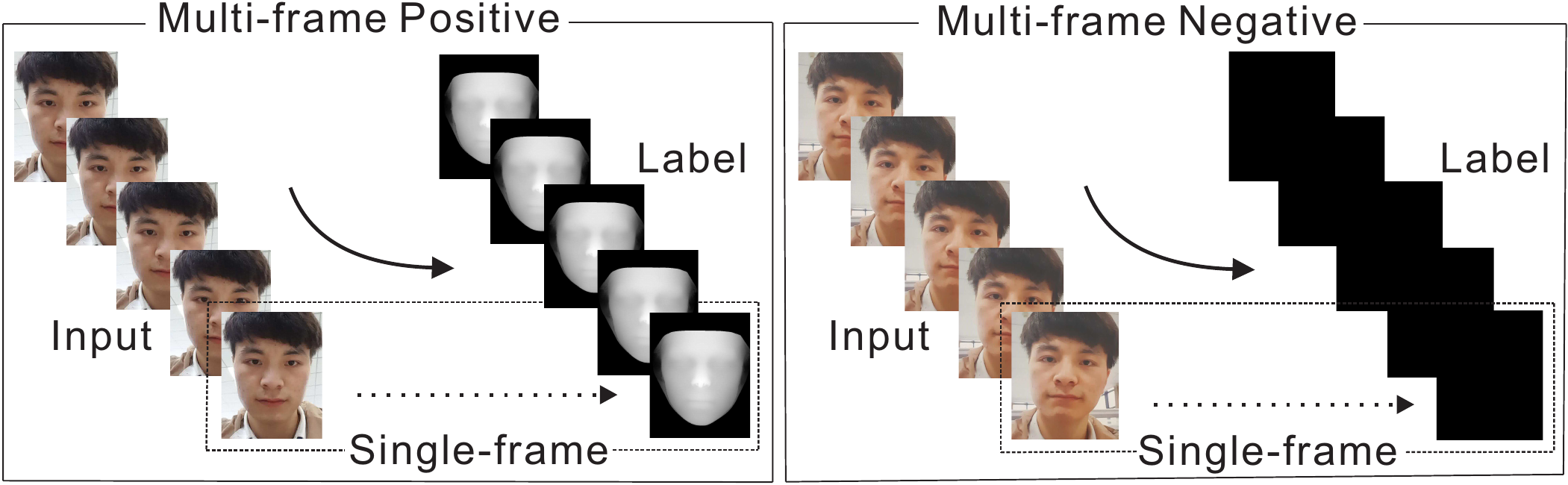}
  \caption{The examples of our input and label for single-frame and multi-frame.
  }
  \label{fig:input_label}
\vspace{-10pt}
\end{figure}
We train the single-frame network with facial depth map, which reveals face location and 3D shape of the face in the 2D plane image, and extracts useful information in the recognition of real face or not. To distinguish living faces from spoofing faces, we normalize living depth map in a range of $[0, 1]$, while setting spoofing depth map to 0~\cite{Liu2018Learning}. %The definition of our depth map is originally beneficial to the testing inference by meaning the facial depth as living score. Simultaneously, we emphasize the essentiality of depth supervision again, that is, depth map provides pixel-wise supervision.

In this module, we adopt dense face alignment method PRNet~\cite{Feng2018Joint} to estimate the 3D shape of the living face. PRNet can be used to project the 3D shape of a complete face into UV space. Through this method, we obtain a group of vertices $V_{n\times3}$ representing $n$ 3D coordinates of facial keypoints. Since these coordinates are sparse when mapped to the 2D plane image, we interpolate them to obtain dense face coordinates. By mapping and normalizing the interpolated coordinates to a 2D plane  image, the generated facial depth map can be represented by $\textbf{\rm{D}} \in \mathbb{R}^{32 \times 32}$. Fig.~\ref{fig:input_label} shows the labels and corresponding inputs of the proposed model.
%Since these coordinates are sparse when mapped to the plane 2D image, we tend to implement interpolation on $V_{n\times3}$ for dense face coordinates. As such, we specially calculate the min/max values of $x$ and $y$ in $V_{n\times3}$, and then build a meshgrid based on them. Afterwards, we interpolate the meshgrid for the smooth $z$ value. By mapping the interpolated $V_{n'\times3}$ to a plane 2D image, and normalizing the face depth map to $[0, 1]$ by minimal and maximal $z$ values, the generated facial depth map can be finally represented by $\textbf{\rm{D}} \in \mathbb{R}^{32 \times 32}$. Figure~\ref{fig:input_label} shows the labels and corresponding inputs of our proposed model.
\subsubsection{Network Structure}
\label{sec:methods}

\begin{figure}[t]
\setlength{\belowcaptionskip}{-0.1cm}
  \centering
  \includegraphics[width=0.40\textwidth]{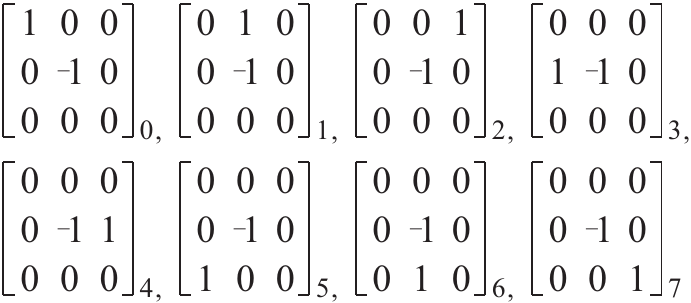}\\
  \caption{The kernel $\textbf{\rm{K}}_{i}^{contrast}$ in contrastive depth loss.}
  \label{fig:K_contrast}
\vspace{-5pt}
\end{figure}

\begin{figure*}[!htb]
  \centering
  \includegraphics[width=1.0\textwidth]{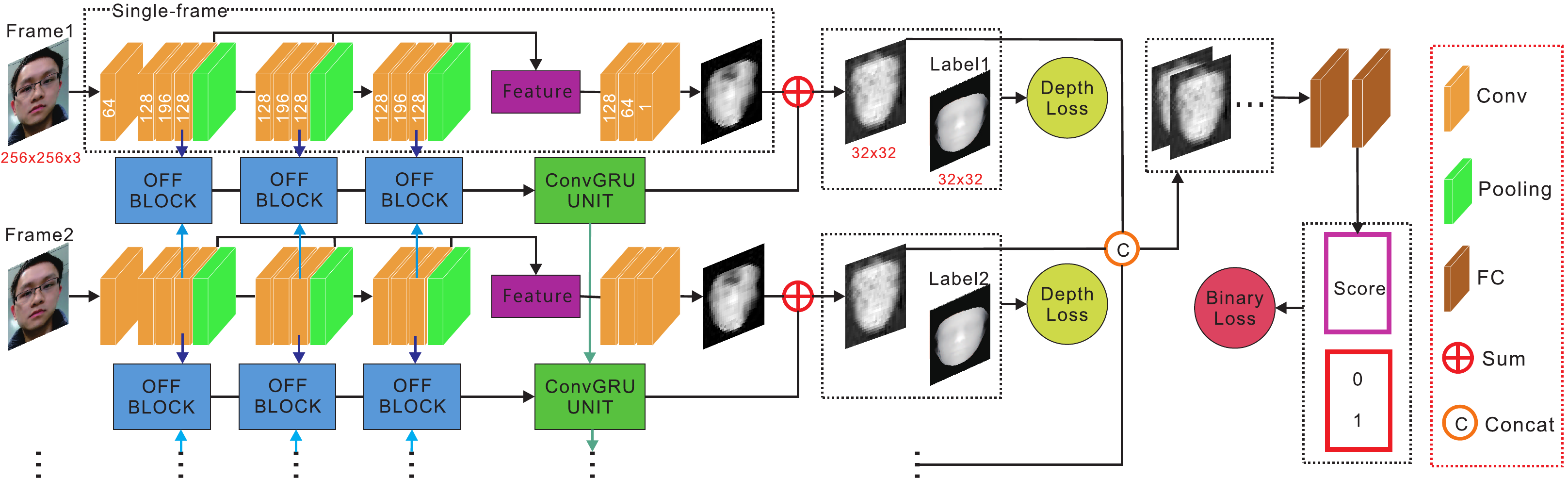}%{./figures/figure_2_pipeline.eps}
  \caption{\textbf{The pipeline of proposed architecture.} The inputs are consecutive frames in a fixed interval. Our single-frame part aims to extract features at various levels and to output the single-frame estimated facial depth. OFF blocks take single-frame features from two consecutive frames as inputs and calculate short-term motion features. Then the final OFF features are fed into the ConvGRUs to obtain long-term motion information, and output the residual of single-frame facial depth. Finally, the combined estimated multi-frame depth maps are supervised by the depth loss and binary loss in respective manners.}
  \label{fig:pipeline}
\vspace{-10pt}
\end{figure*}

As shown in Fig.~\ref{fig:pipeline}, there are three main cascaded blocks connected after one convolution layer in our single-frame part. Each block is composed of three convolution layers and one pooling layer. We resize pooling features to the pre-defined size of $32 \times 32$ and concatenate them into one tensor, which is used to regress the depth map followed by three subsequent convolutional groups.

Given an original RGB image $\rm{I}^{256 \times 256 \times 3}$, we can obtain the estimated depth map $\textbf{\rm{D}}_{single} \in \mathbb{R}^{32 \times 32}$ from the single-frame part. Supervised by the generated ``ground truth'' depth $\textbf{\rm{D}} \in \mathbb{R}^{32 \times 32}$, we design a novel depth loss with the following two parts. One is squared Euclidean norm loss between $\textbf{\rm{D}}_{single}$ and $\textbf{\rm{D}}$ for an absolute depth regression:
\begin{equation}
{\rm{L}}_{single}^{absolute} = ||\textbf{{\rm{D}}}_{single} - \textbf{{\rm{D}}}||_{2}^{2},
\label{eq:single_absolute_loss}
\end{equation}
and the other is a contrastive depth loss:
\begin{equation}
\small
\begin{split}
{\rm{L}}_{single}^{contrast} = & \sum_i{||\textbf{\rm{K}}_{i}^{contrast} \odot \textbf{\rm{D}}_{single} - \textbf{\rm{K}}_{i}^{contrast} \odot \textbf{\rm{D}}||_{2}^{2}},
\label{eq:single_contrast_loss}
\end{split}
\end{equation}
where $\odot$ represents depthwise separable convolution operation~\cite{Howard2017MobileNets}, $\textbf{\rm{K}}_{i}^{contrast}$ represents the contrastive convolution kernel shown in Fig.~\ref{fig:K_contrast}, and $i$ indexes the location of ``1" around ``-1". As shown in Eq.~\ref{eq:single_contrast_loss} and Fig.~\ref{fig:K_contrast}, the contrastive depth loss aims to learn the topography of each pixel, which gives constraints to the contrast from the pixel to its neighbors.

Putting together, the single-frame loss can be written as:
\begin{equation}
{\rm{L}}_{single} = {\rm{L}}_{single}^{absolute} + {\rm{L}}_{single}^{contrast},
\label{eq:single_total_loss}
\end{equation}
where ${\rm{L}}_{single}$ is the final loss used in our single-frame part.
%where we set equal ratio of absolute and contrastive depth loss.
%Driven by the depth loss, our single-frame model can learn more informative features.
\subsection{Multi-frame Part}
\label{subsec:Multi-frame-Computation}
In this part, we expore the short-term and the long-term motion in temporal domain. Specially, the short-term motion is extracted by optical flow guided feature block (OFFB) and the long-term motion is obtained by ConvGRU module.

\subsubsection{Short-term Motion}
\begin{figure}[htbp]
  \includegraphics[width=0.48\textwidth]{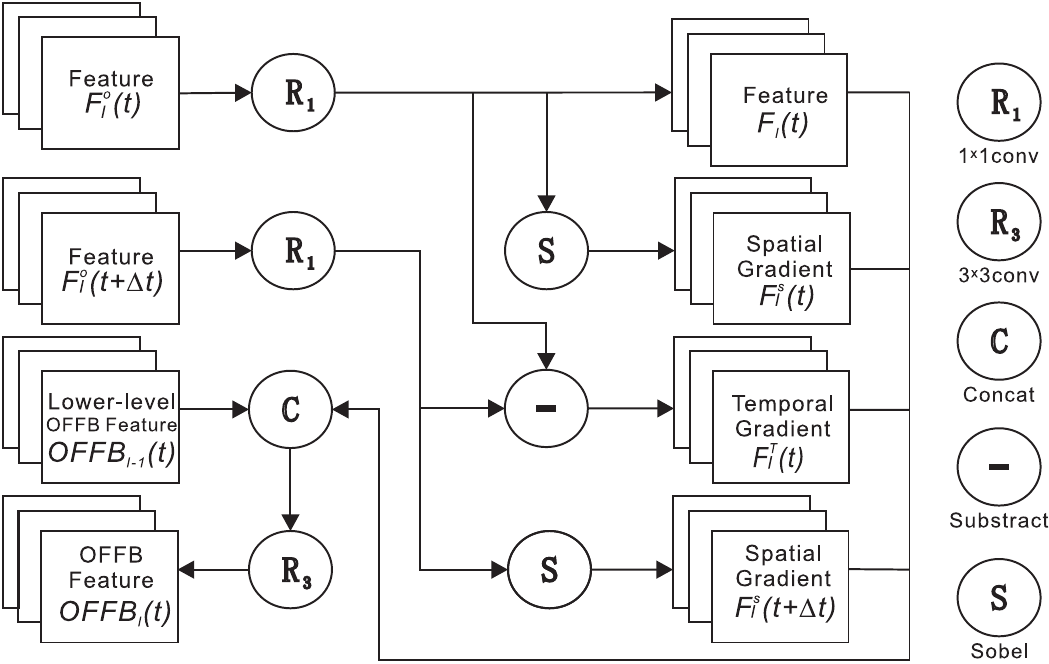}
  \caption{{The architecture of our OFF block.} }
  \label{fig:OFF-BLOCK}
\vspace{-10pt}
\end{figure}
We use OFF block (OFFB) derived from ~\cite{Sun2018Optical} to extract short-term motion. The famous brightness constant constraint in traditional optical flow can be formulated as:
\begin{equation}
I(x,y,t) = I(x+\triangle x, y+\triangle y, t+\triangle t),
\label{eq:bcc}
\end{equation}
where $I(x,y,t)$ represents the brightness at the location $(x,y)$ of the frame at time $t$. %The equation denotes that when a pixel at $(x,y)$ moves $(\triangle x, \triangle y)$ in time span of $\triangle t$, the brightness doesn't change.
%According to Taylor formula, equation~\ref{eq:bcc} can be derived to:
%\begin{equation}
%\frac{\partial I(p)}{\partial x} \triangle x + \frac{\partial I(p)}{\partial y} \triangle y + \frac{\partial I(p)}{\partial t} \triangle t = 0,
%\label{eq:bcc1}
%\end{equation}
From Eq.~\ref{eq:bcc}, we can derive:
\begin{equation}
\frac{\partial I(p)}{\partial x} v_x + \frac{\partial I(p)}{\partial y} v_y + \frac{\partial I(p)}{\partial t} = 0,
\label{eq:bcc2}
\end{equation}
where $p = (x,y,t)$, $v_x$ and $v_y$ represent the two dimensional velocity of the pixel at $p$. $\frac{\partial I(p)}{\partial x}$, $\frac{\partial I(p)}{\partial y}$ and $\frac{\partial I(p)}{\partial t}$ are gradients of $I(p)$ with respects to $x$, $y$ and the time, respectively. $(v_x, v_y)$ is exactly the optical flow. In Eq.~\ref{eq:bcc2}, we can see that $\vec{I}(p) = [\frac{\partial I(p)}{\partial x}, \frac{\partial I(p)}{\partial y}, \frac{\partial I(p)}{\partial t}]$ is orthogonal to $[v_x, v_y, 1]$. Obviously, $\vec{I}(p)$ is guided by the optical flow. By replacing image $I$  by features level in equations above, $\vec{I}(p)$ can be reformulated as:
%\begin{small}
\begin{equation}
\begin{split}
\vec{F}(I;w;t)(p) =&
[\frac{\partial F(I;w;t)(p)}{\partial x}, \frac{\partial F(I;w;t)(p)}{\partial y}, \\
& \frac{\partial F(I;w;t)(p)}{\partial t}],
\label{eq:bcc3}
\end{split}
\end{equation}
%\end{small}
where $F$ is the extracted feature from image $I$, and $w$ is the parameters of the feature extraction function. $\vec{F}(I;w;t)(p)$ is called optical flow guided features (OFF)~\cite{Sun2018Optical}, which encodes spatial and temporal gradients guided by the feature-level optical flow.

%For the spatial gradient $F_l^S(t) = [F_l^x(t),F_l^y(t)]$, we can attain $F_l^x(t)$ and $F_l^y(t)$ by applying Sobel operator:
%\begin{align}
%F_l^x(t) &=
%\left[\begin{matrix}
%-1 & 0 & 1 \\
%-2 & 0 & 2 \\
%-1 & 0 & 1
%\end{matrix}\right] \odot F_l(t), \nonumber\\
%F_l^y(t) &=
%\left[\begin{matrix}
%1 & 2 & 1 \\
%0 & 0 & 0 \\
%-1 & -2 & -1
%\end{matrix}\right] \odot F_l(t).
%\end{align}
%The temporal gradient $F_l^T(t)$ is obtained by element-wise operator as follows:
%\begin{equation}
%F_l^T(t) = F_l(t+\triangle t) - F_l(t).
%\end{equation}

Different from OFF~\cite{Sun2018Optical}, we add extra spatial and temporal gradients and morphological information into our own OFF block (OFFB). As shown in Fig.~\ref{fig:OFF-BLOCK}, there are five sub-modules: $F_l(t), F_l^S(t)$, $F_l^S(t+\triangle t)$, $F_l^T(t)$, $OFFB_{l-1}(t)$, where $F_l(t)$ is reduced from the original single-frame feature $F^O_l(t)$ at feature level $l$ and time $t$ to get foundamental morphological information, $F_l^S(t)$ and $F_l^S(t+\triangle t)$ are the spatial gradients at time $t$ and $t+\triangle t$, $F_l^T(t)$ is the temporal gradient between time $t$ and $t+\triangle t$, and $OFFB_{l-1}(t)$ is the OFFB feature from the previous feature level $l-1$. To reduce the training burden, we introduce a reduction operator -- $1\times 1$ convolution layer -- at the original feature. We concatenate all of the five sub-modules, and feed them into a $3\times 3$ convolution layer to reduce the feature dimension and establish $OFFB_{l}(t)$. More details about spatial gradient $F_l^S(t)$ and temporal gradient $F_l^T(t)$ can refer to ~\cite{Sun2018Optical}. In the following of this paper, we denote the output of the final OFF block by $OFFB(t)$.

\subsubsection{Long-term Motion}
As a short-term motion feature, OFFB feature primarily captures the motion information between two consecutive frames, whereas has difficulty in fusing long-sequence motion. In this regard, we resort to ConvGRU module to gain the long-term motion.

GRU~\cite{Cho2014Learning} derives from Long Short-term Memory (LSTM)~\cite{Graves1997Long}, which has simpler structure and fewer trainable parameters. Similar to LSTM, GRU aims to process long sequence information as well.
%This operation is similar to ConvLSTM~\cite{Shi2015Convolutional}, which utilizes convolution layer in LSTM to estimate precipitation nowcasting. In the light of the better performance of GRU on smaller dataset~\cite{Chung2014Empirical}, we decide to implement ConvGRU in our work.

However, normal GRU neglects the spatial information in its hidden units. Thus, we should take into account the convolution operation in the hidden layers to model the spatiotemporal sequences, and thus propose the Convolution Gated Recurrent Units (ConvGRU) related to ConvLSTM~\cite{Shi2015Convolutional}. The ConvGRU can be described below:
\begin{align}
R_t &= \sigma(K_r \otimes [H_{t-1}, X_t]),\nonumber\\
U_t &= \sigma(K_u \otimes [H_{t-1}, X_t]),\nonumber\\
\hat{H}_t &= \tanh(K_{\hat{h}} \otimes [R_t * H_{t-1}, X_t]),\nonumber\\
H_t &= (1 - U_t)*H_{t-1} + U_t * \hat{H}_t,
\label{eq:ConvGRU}
\end{align}
where $X_t, H_t, U_t$ and $R_t$ are the matrix of input, output, update gate and reset gate, $K_r, K_u, K_{\hat{h}}$ are the kernels in the convolution layer, $\otimes$ is convolution operation, $*$ denotes element wise product, and $\sigma$ is the sigmoid activation function. By feeding $\{OFFB(t)\}^{N_f-1}$ into $\{X_t\}^{N_f-1}$, we set up depth maps $\{\textbf{\rm{D}}_{multi}^{t}\}^{N_f-1}$, where $\textbf{\rm{D}}_{multi}^{t} = H_t$ and $N_f$ denotes the number of input frames. Then, based on the residual idea, we integrate the single depth map and multi depth map:
\begin{equation}
\textbf{\rm{D}}_{fusion}^{t} = \alpha \cdot \textbf{\rm{D}}_{single}^{t} + (1-\alpha) \cdot \textbf{\rm{D}}_{multi}^{t},
\alpha \in [0,1]
\end{equation}
where $\alpha$ is the weight of $\textbf{\rm{D}}_{single}^{t}$ in $\textbf{\rm{D}}_{fusion}^{t}$. Finally, we build up the set of multi-frame depth maps $\{\textbf{\rm{D}}_{fusion}^t\}^{N_f - 1}$.

\subsubsection{Multi-frame Loss}
We make the final decision to discriminate live vs. spoof in the multi-frame part. Nevertheless, in view of the potentially unclear depth map, we hereby consider a binary loss when looking for the difference between living and spoofing depth map. Note that the depth supervision is decisive, whereas the binary supervision takes an assistant role to discriminate the different kinds of depth maps. In this regard, we establish the following multi-frame loss:
\begin{equation}
{\rm{L}}_{multi}^{absolute}(t) = ||\textbf{\rm{D}}_{fusion}^t - \textbf{\rm{D}}^t||_{2}^{2},
\end{equation}

\begin{equation}
\begin{split}
{\rm{L}}_{multi}^{contrast}(t) = &  \sum_i{||\textbf{\rm{K}}_{i}^{contrast} \odot \textbf{\rm{D}}_{fusion}^t -}
\\ &  \textbf{\rm{K}}_{i}^{contrast} \odot   \textbf{\rm{D}}^t||_{2}^{2},
\end{split}
\end{equation}

\begin{equation}
{\rm{L}}_{multi}^{depth} = \sum^{N_f-1}_t{({\rm{L}}_{multi}^{absolute}(t) + {\rm{L}}_{multi}^{contrast}(t))},
\end{equation}

\begin{equation}
{\rm{L}}_{multi}^{binary} = -\textbf{{\rm{B}}}^t * log(fcs(  [\{\textbf{{\rm{D}}}_{fusion}^t\}^{N_f - 1}]  )),
\label{eq:binary_loss}
\end{equation}

\begin{equation}
{{\rm{L}}_{multi}} = \beta \cdot {{\rm{L}}_{multi}^{binary}} + (1-\beta) \cdot {\rm{L}}_{multi}^{depth},
\label{eq:mutil_loss}
\end{equation}
where $\textbf{{\rm{D}}}^{t}$ and ${\rm{B}}^t$ are depth label and binary label at time $t$, $[\{\textbf{{\rm{D}}}_{fusion}^t\}^{N_f - 1}]$ is the concatenated depth maps of $N_f-1$ frames, $fcs$ denotes two fully connected layers and one softmax layer after the concatenated depth maps, which outputs the logits of two classes, $\beta$ is the weight of binary loss in the final multi-frame loss ${{\rm{L}}_{multi}}$. In Eq.~\ref{eq:binary_loss}, we use cross entropy loss to calculate the binary loss. In Eq.~\ref{eq:mutil_loss}, we simply take a sum of the binary loss and depth loss.

\section{Experiment}
\subsection{Databases and Metrics}
\subsubsection{Databases}
Four databases - OULU-NPU~\cite{Boulkenafet2017OULU}, ~SiW~\cite{Liu2018Learning}, CASIA-MFSD~\cite{Zhang2012A}, Replay-Attack~\cite{Chingovska2012On} are used in our experiment. OULU-NPU~\cite{Boulkenafet2017OULU} is a high-resolution database, consisting of 4950 real access and spoofing videos. This database contains four protocols to validate the generalization of models. ~SiW~\cite{Liu2018Learning} contains more live subjects and three protocols are used for testing. CASIA-MFSD~\cite{Zhang2012A} and Replay-Attack~\cite{Chingovska2012On} are databases which contain low-resolution videos. We use these two databases for cross testing.
%The ratio of real videos to attack videos in OULU-NPU~\cite{Boulkenafet2017OULU} is 1:4.
\subsubsection{Metrics}
In OULU-NPU and SiW dataset, we follow the original protocols and metrics for a fair comparison. OULU-NPU and SiW utilize 1) Attack Presentation
Classification Error Rate $APCER$, which evaluates the highest error among all PAIs(e.g.
print or display), 2) Bona Fide Presentation Classification Error Rate $BPCER$, which evaluates error of real access data, and 3) $ACER$~\cite{ACER}, which evaluates the mean of $APCER$ and $BPCER$:
\begin{equation}
ACER=\frac{APCER+BPCER}{2}.
\end{equation}
HTER is adopted in the cross testing between CASIA-MFSD and Repaly-Attack, which evaluates the mean of False Rejection Rate (FRR) and the False Acceptance Rate (FAR):
\begin{equation}
HTER=\frac{FRR+FAR}{2}.
\end{equation}
\subsection{Implementation Details}
\subsubsection{Training Strategy}
The proposed method combines the single-frame part and multi-frame part.
Two-stage strategy is applied in the training process. \emph{Stage 1:} We train the single-frame part by the single-frame depth loss, in order to learn a fundamental representation. \emph{Stage 2:} We fix the parameters of the single-frame part, and finetune the parameters of multi-frame part by the depth loss and binary loss. Note that the diverse data should be adequately shuffled for the stability of training and generalization of the learnt model. The network is fed by $N_f$ frames, which are sampled by a interval of three frames. This sampling interval makes sampled frames maintain enough temporal information in the limitation of GPU memory.
\subsubsection{Testing Strategy}
For the final classification score, we feed the sequential frames into the network and obtain depth maps ${\{\textbf{\rm{D}}_{fusion}^t\}}^{N_f-1}$ and the living logits $\hat{b}$ in $fcs(\textbf{\rm{D}}_{fusion}^t)$. The final living score can be obtained by:
\begin{equation}
score = \beta \cdot \hat{b} + (1-\beta) \cdot
\frac{ \sum_t^{N_f-1}{||\textbf{\rm{D}}_{fusion}^t * \textbf{\rm{M}}_{fusion}^t||_1} }{ N_f-1 },
\label{eq:testing_inference}
\end{equation}
where $\beta$ is the same as that in equation~\ref{eq:mutil_loss}, $\textbf{\rm{M}}_{fusion}^t$ is the mask of face at frame $t$ , which can be generated by the dense face landmarks in PRNet~\cite{Feng2018Joint}, and the second module denotes that we compute the mean of depth values in the facial areas as one part of the score.
\subsubsection{Hyperparameter Setting}
We implement our proposed method in Tensorflow~\cite{Abadi2016TensorFlow}, with a learning rate of 3e-3 for single-frame part and 1e-2 for multi-frame part. The batch size of single-frame part is 10 and that of multi-frame part is 2 with $N_f$ being 5 in most of our experiment, except that batch size being 4 and $N_f$ being 3 in protocol 2 of OULU-NPU. Adadelta optimizer is used in our training procedure, with $\rho$ as 0.95 and $\epsilon$ as 1e-8. We set $\alpha$ and $\beta$ to optimal values by our experimental experience, and according to the anylysis of the following section that protocol 4 in OULU-NPU is most challenging to the generalization, we recommend that the parameters $\alpha=0.8$ and $\beta=0.9$ are suitable for the realistic scenes.
%we set $(\alpha, \beta)$ to $ (0.01, 0.01), (0.01, 0.99), (0.8, 0.9)$ and $(0.8, 0.9)$ in protocol 1,2,3,4 of OULU-NPU respectively. In CASIA-MFSD and Replay-Attack cross-testing, $(\alpha, \beta)$ is set to (0.8, 0.9) from CASIA-MFSD to Replay-Attack and (0.99, 0.01) from Replay-Attack to CASIA-MFSD.
\subsection{Experimental Comparison}
\subsubsection{Intra Testing}
\label{sec:intra_testing}
\begin{table}
\resizebox{0.48\textwidth}{!}{
\begin{tabular}{|c|c|c|c|c|}%一个c表示有一列\emph{}，格式为居中显示(center)
\hline
Prot. & Method & APCER(\%) & BPCER(\%) & ACER(\%) \\
\hline
\multirow{4}{*}{1} &CPqD &2.9 &10.8 & 6.9 \\
       \cline{2-5} &GRADIANT &1.3 &12.5 & 6.9 \\
       \cline{2-5} &FAS-BAS~\cite{Liu2018Learning} &1.6 &1.6 & 1.6 \\
       \cline{2-5} &\textbf{OURs} &2.5 &0.0 & \textbf{1.3} \\
\hline
\multirow{4}{*}{2} &MixedFASNet &9.7 &2.5 & 6.1 \\
       \cline{2-5} &FAS-BAS~\cite{Liu2018Learning} &2.7 &2.7 & 2.7 \\
       \cline{2-5} &GRADIANT &3.1 &1.9 & 2.5 \\
       \cline{2-5} &\textbf{OURs} &1.7 &2.0 & \textbf{1.9} \\
\hline
\multirow{4}{*}{3} &MixedFASNet &5.3$\pm$6.7 &7.8$\pm$5.5 &6.5$\pm$4.6 \\
       \cline{2-5} &\textbf{OURs} &5.9$\pm$1.9 &5.9$\pm$3.0 &5.9$\pm$\textbf{1.0} \\
       \cline{2-5} &GRADIANT &2.6$\pm$3.9 &5.0$\pm$5.3 &3.8$\pm$2.4 \\
       \cline{2-5} &FAS-BAS~\cite{Liu2018Learning} &2.7$\pm$1.3 &3.1$\pm$1.7 &\textbf{2.9}$\pm$1.5 \\
\hline
\multirow{4}{*}{4} &Massy\_HNU &35.8$\pm$35.3 &8.3$\pm$4.1 &22.1$\pm$17.6 \\
       \cline{2-5} &GRADIANT &5.0$\pm$4.5 &15.0$\pm$7.1 &10.0$\pm$5.0 \\
       \cline{2-5} &FAS-BAS~\cite{Liu2018Learning} &9.3$\pm$5.6 &10.4$\pm$6.0 &9.5$\pm$6.0 \\
       \cline{2-5} &\textbf{OURs} &14.2$\pm$8.7 &4.2$\pm$3.8 &\textbf{9.2$\pm$3.4} \\
\hline
\end{tabular}
}%\resizebox{\textwidth}{!}{
\caption{The results of intra testing on four protocols of OULU-NPU~\cite{Boulkenafet2017OULU}.} %FAS-BAS is the model in ~\cite{Liu2018Learning}}
\label{tab:OULU}
\end{table}

\begin{table}
\resizebox{0.48\textwidth}{!}{
\begin{tabular}{|c|c|c|c|c|}%一个c表示有一列\emph{}，格式为居中显示(center)
\hline
Prot. & Method & APCER(\%) & BPCER(\%) & ACER(\%) \\
\hline
\multirow{2}{*}{1} &FAS-BAS~\cite{Liu2018Learning} &3.58 &3.58 &3.58 \\
       \cline{2-5} &\textbf{OURs} &0.96 &0.50 &\textbf{0.73} \\
\hline
\multirow{2}{*}{2} &FAS-BAS~\cite{Liu2018Learning} &0.57$\pm$0.69 &0.57$\pm$0.69 &0.57$\pm$0.69 \\
       \cline{2-5} &\textbf{OURs} &0.08$\pm$0.17 &0.21$\pm$0.16 &\textbf{0.15$\pm$0.14} \\
\hline
\multirow{2}{*}{3} &FAS-BAS~\cite{Liu2018Learning} &8.31$\pm$3.81 &8.31$\pm$3.80 &8.31$\pm$3.81 \\
       \cline{2-5} &\textbf{OURs} &3.10$\pm$0.79 &3.09$\pm$0.83 &\textbf{3.10$\pm$0.81} \\
\hline
\end{tabular}
}%\resizebox{\textwidth}{!}{
\caption{The results of intra testing on three protocols of SiW~\cite{Liu2018Learning}. } %FAS-BAS is the model in ~\cite{Liu2018Learning}}
\label{tab:SiW}
%\vspace{-5pt}
\end{table}

We compare the performance of intra testing on OULU-NPU and SiW dataset. OULU-NPU proposes four protocols to evaluate the generalization of the developed face presentation attack detection (PAD) methods. Protocol 1 is designed to evaluate the generalization of PAD methods under previously unseen illumination and background scene. Protocol 2 is designed to evaluate the generalization of PAD methods under unseen attack medium, such as unseen printers or displays. Protocol 3 utilizes a Leave One Camera Out (LOCO) protocol, in order to study the effect of the input camera variation. Protocol 4 considers all above factors and integrates all the constraints from protocols 1 to 3, so protocol 4 is the most challenging. Table~\ref{tab:OULU} shows that our proposed method ranks first on three protocols - protocol 1,2,4, and ranks third on protocol 3. We can see that our model performs well at the generalization of external environment and attack mediums, and is slightly worse when it comes to the input camera variation.
It's worth noting that our proposed method has the lowest mean and std of ACER in protocol 4, which is most suitable for the real-life scenarios.

Tab.~\ref{tab:SiW} compares the performance of our method with FAS-BAS~\cite{Liu2018Learning} on SiW~\cite{Liu2018Learning}. According to the purposes of three protocols on SiW~\cite{Liu2018Learning} and the results in Tab.~\ref{tab:SiW}, we can see that our method performs great advantages on the generalization of (1) variations of face pose and expression, (2) variations of different spoof mediums, (3) cross PAI.

\vspace{-10pt}
\subsubsection{Cross Testing}
%\noindent
\newcommand{\tabincell}[2]{\begin{tabular}{@{}#1@{}}#2\end{tabular}}
\begin{table}
\resizebox{0.48\textwidth}{!}{
\begin{tabular}{|c|c|c|c|c|}%一个c表示有一列\emph{}，格式为居中显示(center)
\hline
\multirow{2}{*}{Method} &Train &Test &Train &Test\\
\cline{2-5} &\tabincell{c}{CASIA-\\MFSD} &\tabincell{c}{Repaly-\\Attack} &\tabincell{c}{Repaly-\\Attack} &\tabincell{c}{CASIA-\\MFSD}\\
\hline
Motion~\cite{Pereira2013Can}
&\multicolumn{2}{c|}{50.2} &\multicolumn{2}{c|}{47.9} \\
\hline
LBP-1~\cite{Pereira2013Can}
&\multicolumn{2}{c|}{55.9} &\multicolumn{2}{c|}{57.6} \\
\hline
LBP-TOP~\cite{Pereira2013Can}
&\multicolumn{2}{c|}{49.7} &\multicolumn{2}{c|}{60.6} \\
\hline
Motion-Mag~\cite{bharadwaj2013computationally}
&\multicolumn{2}{c|}{50.1} &\multicolumn{2}{c|}{47.0} \\
\hline
Spectral cubes~\cite{pinto2015face}
&\multicolumn{2}{c|}{34.4} &\multicolumn{2}{c|}{50.0} \\
\hline
CNN~\cite{Yang2014Learn}
&\multicolumn{2}{c|}{48.5} &\multicolumn{2}{c|}{45.5} \\
\hline
LBP-2~\cite{boulkenafet2015face}
&\multicolumn{2}{c|}{47.0} &\multicolumn{2}{c|}{39.6} \\
\hline
Colour Texture~\cite{Boulkenafet2017Face}
&\multicolumn{2}{c|}{30.3} &\multicolumn{2}{c|}{37.7} \\
\hline
FAS-BAS~\cite{Liu2018Learning}
&\multicolumn{2}{c|}{27.6} &\multicolumn{2}{c|}{28.4} \\
\hline
\textbf{OURs}
&\multicolumn{2}{c|}{\textbf{17.5}} &\multicolumn{2}{c|}{\textbf{24.0}} \\
\hline
\end{tabular}
}%\resizebox{\textwidth}{!}{
\caption{The results of cross testing between CASIA-MFSD and Replay-Attack. The evaluation metric is HTER(\%).}
\label{tab:cross-testing}
\vspace{-15pt}
\end{table}
We utilize CASIA-MFSD and Replay-Attack dataset to perform cross testing. This can be regarded as two testing protocols. One is training on the CASIA-MFSD and testing on Replay-Attack, which we name protocol CR; the other is training on the Replay-Attack and testing on CASIA-MFSD, which we name protocol RC. In table~\ref{tab:cross-testing}, we see HTER of our proposed method is 17.5 on protocol CR and 24.0 on protocol RC, reducing 36.6\% and 15.5\% respectively compared with the previous state of the art. The improvement of performance on cross testing demonstrates the generalization and superiority of our proposed method.

\subsubsection{Ablation Study}
\begin{table}[t]
\resizebox{0.48\textwidth}{!}{
\begin{tabular}{|l|c c c c c|c|}%一个c表示有一列\emph{}，格式为居中显示(center)
\hline
Module &Single-frame &OFF-block &ConvGRU &D-S &B-S &ACER(\%)\\
\hline
Model 1 &$\surd$ &\   &\  &$\surd$  &\  &4.4 \\
Model 2$^{*}$ &$\surd$ &$\surd$ &\  &$\surd$  &\  &3.5 \\
Model 2 &$\surd$ &$\surd$ &\  &$\surd$  &$\surd$  &2.3 \\
Model 3$^{*}$ &$\surd$ &\ &$\surd$  &$\surd$  &\  &3.3 \\
Model 3 &$\surd$ &\ &$\surd$  &$\surd$  &$\surd$  &3.1 \\
Model 4$^{*}$ &$\surd$ &$\surd$ &$\surd$  &$\surd$  &\ &1.7 \\
Model 4 &$\surd$ &$\surd$ &$\surd$  &$\surd$  &$\surd$ &\textbf{1.3} \\
\hline
\end{tabular}
}%\resizebox{\textwidth}{!}{
\caption{The results of ablation study on protocol 1 of OULU-NPU. B-S denotes binary-supervision, and D-S denotes depth-supervision.}
\label{tab:ablation}
\end{table}

\begin{table}[t]
\resizebox{0.48\textwidth}{!}{
\begin{tabular}{|l|c c|c|}%一个c表示有一列\emph{}，格式为居中显示(center)
\hline
\multirow{2}{*}{Module} &\multicolumn{2}{c|}{Depth Loss} &\multirow{2}{*}{ACER(\%)}\\
\cline{2-3} &Euclidean Depth Loss &Contrastive Depth Loss  &\\
\hline
Model 1{$^*$}  &$\surd$ & &5.8 \\
%\hline
Model 1  &$\surd$ &$\surd$ &\textbf{4.4} \\
\hline
\end{tabular}
}%\resizebox{\textwidth}{!}{
\caption{The results of ablation study on the influence of contrastive depth loss on protocol 1 of OULU-NPU.}
\label{tab:eucl_contra}
\vspace{-10pt}
\end{table}
%\begin{figure}[t]
%  % Requires \usepackage{graphicx}
%  \includegraphics[width=0.46\textwidth]{./figures/roc_ablation.eps}
%  \caption{
%  The ROC curves for ablation study on protocol 1 of OULU-NPU.
%  }
%  \label{fig:roc_ablation}
%\end{figure}
We implement experiment on seven architectures to demonstrate the advantages of our proposed sequential structure under the supervision of depth. As shown in table~\ref{tab:ablation}, Model 1 is the single-frame part of our method. Model 2 combines single-frame CNN with OFFB under binary supervision and depth supervision. Model 3 combines single-frame CNN with ConvGRU under binary supervision and depth supervision. Model 2$^*$ and model 3$^*$ discard the binary supervision. Model 4 is our complete architecture, integrating all modules. Model 4$^*$ discards binary supervision compared with model 4. Comparing ACER of model 2$^*$ and model 3$^*$ with that of model 1, we see that our OFFB module and ConvGRU module both improve the performance of face anti-spoofing. And the ACER of model 4$^*$ shows that the combination of OFFB module and ConvGRU module has more positive effects. Via discarding the binary supervision, we test the effect of binary supervision on our model 2$^*$, 3$^*$ and 4$^*$. In this model, we find that multi-frame model with simple depth supervision can also outperform the single-frame model and binary supervision indeed assists the model to distinguish live vs. spoof.
%In this model, we find that simple depth supervision is able to assist the sequential model to learn discriminative features and has a lower ACER.

In table~\ref{tab:eucl_contra}, we study the influence of contrastive depth loss. Model 1 is our single-frame model supervised by both the euclidiean depth loss and the contrastive depth loss, while model 1$^{*}$ is supervised only by euclidiean depth loss. Comparing model 1 with model 1$^{*}$, we can see that contrastive depth loss can improve the generalization of our model.

Moreover, the inference of the model 1 costs around 18 ms and that of model 5 costs around 96 ms, which indicates that our method is efficient enough to be applied in reality.
\vspace{-5pt}
\subsubsection{Qualitative Analysis}
\begin{figure}[t]
%\setlength{\belowcaptionskip}{-0.3cm}
  %\centering
  \includegraphics[width=0.45\textwidth]{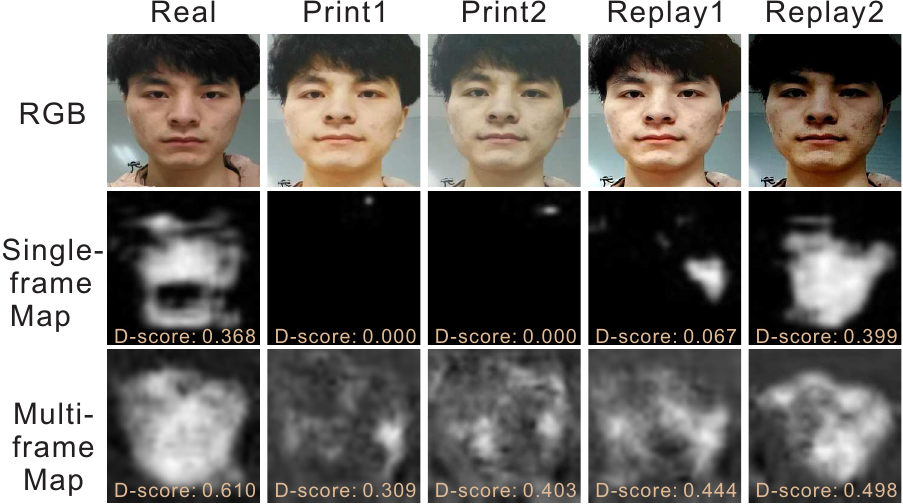}
  \caption{The generated results of a group of hard samples in OULU-NPU. D-score denotes the living score calculated by the mean of depth values in facial area, which is the depth subpart in equation~\ref{eq:testing_inference}.}
  %\caption{The generated results of some samples in OULU-NPU. The three rows represent the original images, single-frame maps and multi-frame maps, respectively. D-score denotes the living score equal to the depth mean of facial area.}
  \label{fig:result_maps}
\vspace{-13pt}
\end{figure}

%\begin{figure}[t]
%  %\centering
%  \includegraphics[width=0.45\textwidth]{./figures/off_v1.eps}
%  \vspace{-10pt}
%  \caption{The visible OFF features of some samples in OULU-NPU. The two rows represent the original images and feature maps, respectively. }
%  \label{fig:result_offs}
%  \vspace{-15pt}
%\end{figure}
Figure~\ref{fig:result_maps} presents the generated depth maps of a group of hard samples for the same person ID in OULU-NPU. From figure~\ref{fig:result_maps}, we can see that our multi-frame maps are more complete than single-frame maps in real scenes. Although the multi-frame maps in spoofing scenes are visually noisier than those in the single-frame maps, the discrimination is obvious when only considering the multi-frame maps themselves. Specially, in single-frame maps, the D-score of real scene is 0.368, which is lower than that of replay2 scene. This is probably caused by the illumination condition in replay2 scene. By contrast, in multi-frame maps, the D-score of real scene is higher than the D-scores in all of the attack scenes. We can observe the discrimination among the multi-frame maps obviously.
%Visually, the multi-frame maps are also more distinguishable between the real scene and attack scenes than single-frame maps.

It's worth noting that this group of images are results of hard samples, which is failure case in single frame but success case in multi frame. We use residual architecture, adding the multi-frame depth to the single-frame depth, which leads to the depth values in multi frame are higher than those of single frame.

\vspace{-5pt}
\section{Discussion}
We only discuss a simple case of equal motion in vertical dimension in Sec.~\ref{subsec:theory}. In reality, the facial variation and motion are more complex, including forward/backward motion, rotation, deformation and so on. In these cases, we still assume that there are discriminative patterns of temporal depth and motion between living and spoofing faces in the temporal domain. Extension of theory and application on these aspects is our future research. Face anti-spoofing based on facial motion and depth is indeed promising and valuable.

\section{Conclusions}
In this paper, we analyze the usefulness of motion and depth in presentation attack detection. According to the analysis, we propose a novel face anti-spoofing method, which is depth supervised and consists of adequate temporal information. To seek the spatiotemporal information, we take OFF block as short-term motion module and ConvGRU as long-term motion module, and then combine them into our architecture. Our proposed method can discover spoof patterns efficiently and accurately under depth-supervision. Extensive experimental results demonstrate the superiority of our method.

%\clearpage
{\small
\bibliographystyle{ieee}
\bibliography{refs}
}

\clearpage
%%%-----------------------------------
%%%%%%%%% TITLE
%\title{Supplementary Material}
%\maketitle
\begin{center}
\Large{\textbf{Appendix}}
\end{center}
%%%%%%%%% BODY TEXT
%\section{Theory}
\begin{appendix}
\section{Temporal Depth in Face Anti-spoofing}
In this section, we use some simple examples to explain that exploiting temporal depth and motion is reasonable in the face anti-spoofing task.
\begin{figure}[htbp]
  \includegraphics[width=0.48\textwidth]{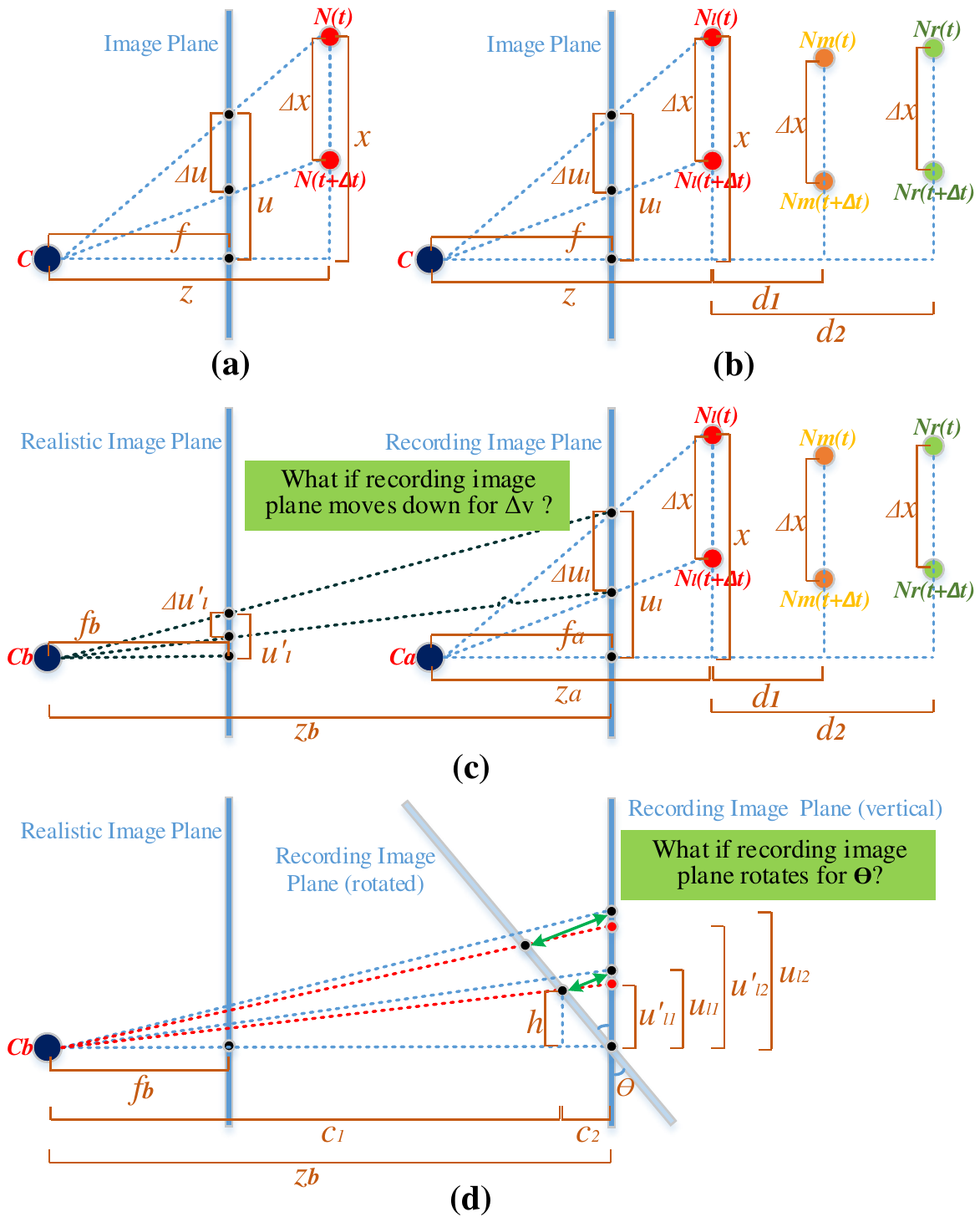}
  \caption{The schematic diagram of motion and depth variation in different scenes.}
  \label{fig:theory1}
\end{figure}

%\subsection{Basic theory}
\subsection{Basic Scene}
As shown in Fig.~\ref{fig:theory1}(a), node $C$ denotes the camera focus. $Image\ Plane$ represents the image plane of camera. $N(t)$ is one facial point at time $t$, and $N(t+\Delta t)$ is the corresponding point when $N(t)$ moves down vertically for $\Delta x$ at time $t+\Delta t$. For example, $N(t)$ can be the point of nose or ear. $f$ denotes the focal distance, and $z$ is the horizontal distance from the focal point to the point $N(t)$. $u$ and $x$ are the corresponding coordinates in vertical dimension. When $N(t)$ moves down vertically to $N(t+\Delta t)$ for $\Delta x$, the motion can be reflected on the image plane as $\Delta u$.
%$f, z, x, \Delta x, u, \Delta u, d$ are corresponding distance.
According to the camera model, we can obtain:
\begin{equation}
\begin{split}
& \frac{x}{u} = \frac{z}{f}, \\
\Leftrightarrow & u = \frac{fx}{z}.
\end{split}
\label{eq:camera_basis}
\end{equation}
When $N(t)$ moves down vertically for $\Delta x$ to $N(t+\Delta t)$, the $\Delta u$ can be achieved:
\begin{equation}
\begin{split}
\Delta u &= \frac{f \Delta x}{z}.
\end{split}
\label{eq:1}
\end{equation}
%where $\Delta u$ is the flow of $N(t)$, which can be calculated by sequential images.

As shown in Fig.~\ref{fig:theory1}(b), to distinguish points $N_l$, $N_m$ and $N_r$, we transform Eq.~\ref{eq:1} and get $\Delta u_l$, $\Delta u_m$ and $\Delta u_r$ ($\Delta u_m$ and $\Delta u_r$ are not shown in the figure):
\begin{equation}
\begin{split}
\Delta u_l &= \frac{f \Delta x}{z},\\
\Delta u_m &= \frac{f \Delta x}{z+d_1}, \\
\Delta u_r &= \frac{f \Delta x}{z+d_2}, \\
\end{split}
\label{eq:2}
\end{equation}
where $d_1$ and $d_2$ are the corresponding depth difference.
From Eq.~\ref{eq:2}, there are:
\begin{equation}
\begin{split}
\frac{\Delta u_l}{\Delta u_m} = \frac{z+d_1}{z} = \frac{d_1}{z} + 1,\\
\frac{\Delta u_l}{\Delta u_r} = \frac{z+d_2}{z} = \frac{d_2}{z} + 1.
\end{split}
\label{eq:3}
\end{equation}
%Considering that $f$ is very little, $\frac{d}{z}$ can be recognized as the ratio of relative depth $d$ to absolute depth $z$. Therefore, depth can be
Removing $z$ from Eq.~\ref{eq:3}, $d_1/d_2$ can be obtained:
\begin{equation}
\begin{split}
\frac{d_1}{d_2} =
\displaystyle{  \frac{  \displaystyle{\frac{\Delta u_l}{\Delta u_m}} - 1}{  \displaystyle{\frac{\Delta u_l}{\Delta u_r}} - 1} },
\end{split}
\label{eq:4}
\end{equation}
In this equation, we can see that the relative depth $d_1/d_2$ can be estimated by the motion of three points, when $d_2 \neq 0$. The equations above are about the real scenes. In the following, we will introduce the derivation of attack scenes.

\subsection{Attack Scene}
\subsubsection{What if the attack carriers move?}
\label{sec:screen_shake}
As shown in Fig.~\ref{fig:theory1}(c), there are two image spaces in attack scenes: one is recording image space, where we replace $z, f$ by $z_a, f_a$, and the other is realistic image space, where we replace $z, f$ by $z_b, f_b$. In the recording image space, it's similar to Eq.~\ref{eq:2}:
\begin{equation}
\begin{split}
\Delta u_l &= \frac{f_a \Delta x}{z_a},\\
\Delta u_m &= \frac{f_a \Delta x}{z_a+d_1}, \\
\Delta u_r &= \frac{f_a \Delta x}{z_a+d_2}, \\
\end{split}
\label{eq:5}
\end{equation}
where $\Delta u_l, \Delta u_m, \Delta u_r$ are the magnitude of optical flow when three points $N_l(t), N_m(t), N_r(t)$ move down vertically for $\Delta x$.

In the realistic image space, there are:
\begin{equation}
\begin{split}
\Delta u'_l &= \frac{f_b \Delta x_l}{z_b},\\
\Delta u'_m &= \frac{f_b \Delta x_m}{z_b}, \\
\Delta u'_r &= \frac{f_b \Delta x_r}{z_b}, \\
\end{split}
\label{eq:6}
\end{equation}
where $\Delta x_l$, $\Delta x_m$ and $\Delta x_r$ are the motion of three points on the recording image plane, and $\Delta u_l, \Delta u_m, \Delta u_r$ are the corresponding values mapping on the realistic image plane.

Actually, there are $\Delta x_l = \Delta u_l, \Delta x_m= \Delta u_m, \Delta x_r= \Delta u_r$, if the recording screen is static. Now, a vertical motion $\Delta v$ is given to the recording screen, just as $\Delta x_l = \Delta u_l + \Delta v, \Delta x_m= \Delta u_m + \Delta v, \Delta x_r= \Delta u_r + \Delta v$. By inserting $\Delta v$, we transform Eq.~\ref{eq:6} into:
\begin{equation}
\begin{split}
\Delta u'_l &= \frac{f_a f_b \Delta x + z_a f_b \Delta v}{z_a z_b},\\
\Delta u'_m &= \frac{f_a f_b \Delta x + ( z_a + d_1 ) f_b \Delta v}{(z_a + d_1) z_b },\\
\Delta u'_r &= \frac{f_a f_b \Delta x + ( z_a + d_2 ) f_b \Delta v}{(z_a + d_2) z_b },\\
\end{split}
\label{eq:7}
\end{equation}
Due to that only $\Delta u'_l, \Delta u'_m, \Delta u'_r$ can be observed directly in the sequential images, we can estimate the relative depth via $\Delta u'_l, \Delta u'_m, \Delta u'_r$. So we leverage Eq.~\ref{eq:4} to estimate the relative depth $d'_1 / d'_2$:
\begin{equation}
\begin{split}
\frac{d'_1}{d'_2} =
\displaystyle{  \frac{  \displaystyle{\frac{\Delta u'_l}{\Delta u'_m}} - 1}{  \displaystyle{\frac{\Delta u'_l}{\Delta u'_r}} - 1} },
\end{split}
\label{eq:extra_1}
\end{equation}
and then we can insert Eq.~\ref{eq:7} into Eq.~\ref{eq:extra_1} to get:
\begin{equation}
\begin{split}
\frac{d'_1}{d'_2} = \frac{d_1}{d_2} \cdot \frac{f_a \Delta x + (z_a + d_2) \Delta v} {f_a \Delta x + (z_a + d_1) \Delta v}.
\end{split}
\label{eq:8}
\end{equation}
%Here, we again note that $d'_1/d'_2$ can be estimated by imitating Eq.~\ref{eq:4}, which indicates that the estimated depth is calculated by three-points flow in the image plane.
According to equations above, some important conclusions can be summarized:

\begin{itemize}
% item 1 print attack
\item If $\Delta x = 0$, the scene can be recognized as print attack and Eq.~\ref{eq:8} will be invalid, for $\Delta u'_l = \Delta u'_r$, and the denominator in Eq.~\ref{eq:extra_1} will be zero. So here we use Eq.~\ref{eq:7} and
\begin{equation}
\begin{split}
\frac{\Delta u'_l}{\Delta u'_m} = \frac{d'_1}{z_b} + 1,\\
\frac{\Delta u'_l}{\Delta u'_r} = \frac{d'_2}{z_b} + 1,
\end{split}
\label{eq:extra_eq3}
\end{equation}
to obtain:
\begin{equation}
\begin{split}
d'_1 = d'_2 = 0.
\end{split}
\label{eq:12}
\end{equation}
In this case, it's obvious that the facial relative depth is abnormal and the face is fake. %and presentation attack will be detected easily.
% item 1 replay attack
\item If $\Delta x \neq 0$, the scene can be recognized as replay attack.
\begin{itemize}
% item 1.1
\item If $\Delta v = 0$, there is:
\begin{equation}
\begin{split}
\frac{d'_1}{d'_2} = \frac{d_1}{d_2}.
\end{split}
\label{eq:9}
\end{equation}
In this case, if these two image planes are parallel and the single-frame model can not detect the static spoof cues, the model will fail in the task of face anti-spoofing, owing to that the model is hard to find the abnormality of relative depth estimated from the facial motion. We call this scene \textbf{Perfect Spoofing Scene(PSS)}. Of course, making up \textbf{PSS} will cost a lot and is approximately impossible in practice.
% item 1.2
\item If $\Delta v \neq 0$ and we want to meet Eq.~\ref{eq:9}, the following equation should be satisfied:
%\begin{equation}
%\begin{split}
%f_a \Delta x + & (z_a + d_2) \Delta v = f_a \Delta x + (z_a + d_1) \Delta v, \\
%& then,\ (d2 - d1) \Delta v = 0.
%\end{split}
%\label{eq:10}
%\end{equation}
\begin{equation}
\begin{split}
\frac{f_a \Delta x + (z_a + d_2) \Delta v} {f_a \Delta x + (z_a + d_1) \Delta v} = 1,
\end{split}
\label{eq:10}
\end{equation}
then,
\begin{equation}
\begin{split}
& (d_2 - d_1) \Delta v  = 0, \\
\Leftrightarrow & d_2 - d_1 = 0, \ if \ \Delta v \neq 0.
\end{split}
\label{eq:10_2}
\end{equation}

However, in our assumption, $d_1 \neq d_2$, so:
\begin{equation}
\begin{split}
\frac{d'_1}{d'_2} \neq \frac{d_1}{d_2}.
\end{split}
\label{eq:11}
\end{equation}
This equation indicates that relative depth can't be estimated preciously, if the attack carrier moves in the replay attack. %And $\Delta v$ usually varies as well as $d'_1/d'_2$ in the long-term sequence.
And $\Delta v$ usually varies when attack carrier moves in the long-term sequence, leading to the variation of $d'_1/d'_2$.
This kind of abnormality is more obvious along with the long-term motion.
\end{itemize}
% item 2 print attack
%Eq.~\ref{eq:7} and Eq.~\ref{eq:3} and
%\begin{equation}
%\begin{split}
%\frac{d'_1}{d'_2} = \frac{d_1}{d_2} \cdot \frac{z_a + d_2}{z_a + d_1}.
%\end{split}
%\label{eq:12}
%\end{equation}
%Due to $d_1 \neq d_2$, the estimated relative depth $d'_1/d'_2$ is not equal to the correct value $d_1/d_2$. Therefore, print attack can be detected by the temporal motion under depth supervision, theoretically. Of course, in real-scene face recognition, real faces are usually motional. So even though when the print carrier is static, just as $\Delta v = 0$ and Eq.~\ref{eq:8} is unusable, the exception in face recognition is obvious, because humans will move their faces more or less in the real scene. That is to say, when the facial motion is absolute none, we can determine this is a spoofing face.

% item 3 depth label
\item If $d_2$ denotes the largest depth difference among facial points, then $d_1/d_2 \in [0, 1]$, showing that constraining depth label of living face to $[0, 1]$ is valid. As analyzed above, for spoofing scenes, the abnormal relative depth usually varies over time, so it is too complex to be computed directly. Therefore, we merely set depth label of spoofing face to all 0 to distinguish it from living label, making the model learn the abnormity under depth supervision itself.

\end{itemize}

\subsubsection{What if the attack carriers rotate?}
As shown in Fig.~\ref{fig:theory1}(d), we rotate the recording image plane for degree $\theta$. $u_{l2}, u_{l1}$ are the coordinates of $N_l(t), N_l(t+\Delta t)$ mapping on the recording image plane.
%The two black points of end of green double arrow segment are corresponding points when the image plane rotates, and the value of $u_{l2}, u_{l1}$ does not change after rotation.
The two black points at the \emph{right} end of green double arrows on recording image plane (vertical) will reach the two black points at the \emph{left} end of green double arrow on recording image plane (rotated), when the recording image plane rotates. And the corresponding values $u_{l2}, u_{l1}$ will not change after rotation.
For convenient computation, we still map the rotated points to the vertical recording image plane. And the coordinates after mapping are $u'_{l2}, u'_{l1}$. $c_1, c_2, h$ are the corresponding distances shown in the figure. According to the relationship of the  foundamental variables, we can obtain:
\begin{equation}
\begin{split}
& h = u_{l1} \cos{\theta}, \\
& \frac{z_b}{c_{1}} = \frac{u'_{l1}}{h}, \\
& c_2 = u_{l1} \sin{\theta}, \\
& c_{1} + c_{2} = z_b.
\end{split}
\label{eq:rotation_1}
\end{equation}
Deriving from equations above, we can get $u'_{l1}$:
\begin{equation}
\begin{split}
u'_{l1} = \frac{z_b u_{l1} \cos{\theta}}{z_b - u_{l1} \sin{\theta}},
\end{split}
\label{eq:rotation_2}
\end{equation}
and $u'_{l2}$ can also be calculated by imitating Eq.~\ref{eq:rotation_2}:
\begin{equation}
\begin{split}
u'_{l2} = \frac{z_b u_{l2} \cos{\theta}}{z_b - u_{l2} \sin{\theta}}.
\end{split}
\label{eq:rotation_3}
\end{equation}
Subtract $u'_{l1}$ from $u'_{l2}$, the following is achieved:
\begin{equation}
\begin{split}
u'_{l2} - u'_{l1} = (u_{l2} - u_{l1}) \cdot \frac{{z_b}^2 \cos{\theta}}{ (z_b - u_{l1} \sin{\theta}) (z_b - u_{l2} \sin{\theta})}.
\end{split}
\label{eq:rotation_4}
\end{equation}
Obviously, $u_{l2} - u_{l1} = \Delta u_l$. We define $u'_{l2} - u'_{l1} = \Delta u^{\theta}_l$. And then we get the following equation:
\begin{equation}
\begin{split}
\Delta u^{\theta}_l &= \Delta u_l \cdot \frac{{z_b}^2 \cos{\theta}}{ (z_b - u_{l1} \sin{\theta}) [z_b - (u_{l1}+\Delta u_l) \sin{\theta}]}, \\
\Delta u^{\theta}_m &= \Delta u_m \cdot \frac{{z_b}^2 \cos{\theta}}{ (z_b - u_{m1} \sin{\theta}) [z_b - (u_{m1}+\Delta u_l) \sin{\theta}]}, \\
\Delta u^{\theta}_r &= \Delta u_r \cdot \frac{{z_b}^2 \cos{\theta}}{ (z_b - u_{r1} \sin{\theta}) [z_b - (u_{r1}+\Delta u_l) \sin{\theta}]},
\end{split}
\label{eq:rotation_5}
\end{equation}
where the relationship between $\Delta u^{\theta}_m, \Delta u^{\theta}_r$ and $N_m(t), N_r(t)$ are just like that between $\Delta u^{\theta}_l$ and $N_l(t)$, as well as $u_{m1}, u_{r1}$. Note that for simplification, we only discuss the situation that $u_{l1}, u_{m1}, u_{r1}$ are all positive.

Reviewing Eq.~\ref{eq:6}, We can confirm that $\Delta x_l = \Delta u^{\theta}_l$, $\Delta x_m = \Delta u^{\theta}_m$, $\Delta x_r = \Delta u^{\theta}_r$. According to Eq.~\ref{eq:extra_1}, the final $d'_1/d'_2$ can be estimated:
\begin{equation}
\begin{split}
\frac{d'_1}{d'_2} =
\displaystyle{  \frac{  \displaystyle{\frac{\Delta u_l}{\Delta u_m}} \cdot \beta_1 - 1}{  \displaystyle{\frac{\Delta u_l}{\Delta u_r}} \cdot \beta_2 - 1} } =
\displaystyle{  \frac{  (\displaystyle{\frac{d_1}{z_a}}+1) \cdot \beta_1 - 1}{  (\displaystyle{\frac{d_2}{z_a}}+1)  \cdot \beta_2 - 1} },
\end{split}
\label{eq:rotation_6}
\end{equation}
where $\beta_1$ and $\beta_2$ can be represented as:
%\begin{equation}
%\begin{split}
%\beta_1 &= \frac{ (z_b - u_{m1}\sin{\theta})[z_b-(u_{m1}+\Delta u_m)\sin{\theta}] }{ (z_b - u_{l1}\sin{\theta})[z_b-(u_{l1}+\Delta u_l)\sin{\theta}] }, \\
%\beta_2 &= \frac{ (z_b - u_{r1}\sin{\theta})[z_b-(u_{r1}+\Delta u_r)\sin{\theta}] }{ (z_b - u_{l1}\sin{\theta})[z_b-(u_{l1}+\Delta u_l)\sin{\theta}] }.
%\end{split}
%\label{eq:rotation_7}
%\end{equation}
\begin{equation}
\begin{split}
\beta_1 &= \frac{ (z_b - u_{m1}\sin{\theta})(z_b-u_{m2}\sin{\theta}) }{ (z_b - u_{l1}\sin{\theta})(z_b-u_{l2}\sin{\theta}) }, \\
\beta_2 &= \frac{ (z_b - u_{r1}\sin{\theta})(z_b-u_{r2}\sin{\theta}) }{ (z_b - u_{l1}\sin{\theta})(z_b-u_{l2}\sin{\theta}) },
\end{split}
\label{eq:rotation_7}
\end{equation}
where $u_{l2} = u_{l1} + \Delta u_l, u_{m2} = u_{m1} + \Delta u_m, u_{r2} = u_{r1} + \Delta u_r$. Observing Eq.~\ref{eq:rotation_6}, we can see if $\beta_1 < 1, \beta_2 > 1$ or $\beta_1 >1, \beta_2 < 1$, there will be $d'_1/d'_2 \neq d_1/d_2$, .

%Now, we will discuss the cases of $\beta_1 < 1, \beta_2 > 1$. In Eq.~\ref{eq:rotation_7}, multipliers are all positive. And when $u_{m1} > u_{l1}$ and $u_{m2} > u_{l2}$, we can get $\beta_1 < 1$. From $u_{m1} > u_{l1}$ and $u_{m2} > u_{l2}$, we can get
Now, we discuss the sufficient condition of $\beta_1 < 1, \beta_2 > 1$. When $u_{m1} > u_{l1}, u_{m2} > u_{l2}, u_{r1} < u_{l1}, u_{r2} < u_{l2}$, the $\beta_1 < 1, \beta_2 > 1$ can be established. Similar to Eq.~\ref{eq:2}, the relationship of variables can be achieved:
\begin{equation}
\begin{split}
& \frac{f_a x_{l1}}{z_a} = u_{l1}, \frac{f_a x_{l2}}{z_a} = u_{l2}, \\
& \frac{f_a x_{m1}}{z_a + d_1} = u_{m1}, \frac{f_a x_{m2}}{z_a + d_1} = u_{m2}, \\
& \frac{f_a x_{r1}}{z_a + d_2} = u_{r1}, \frac{f_a x_{r2}}{z_a + d_2} = u_{r2},
\end{split}
\label{eq:rotation_extra_relation}
\end{equation}
From Eq.~\ref{eq:rotation_extra_relation} and $u_{m1} > u_{l1}, u_{m2} > u_{l2}$, we can obtain:
\begin{equation}
\begin{split}
x_{m1} &> x_{l1} \cdot \frac{z_a+d_1}{z_a}, \\
x_{m1} + \Delta x &> (x_{l1}+\Delta x) \cdot \frac{z_a+d_1}{z_a}, \\
\end{split}
\label{eq:rotation_8}
\end{equation}
\begin{equation}
\begin{split}
x_{r1} &< x_{l1} \cdot \frac{z_a+d_2}{z_a}, \\
x_{r1} + \Delta x &< (x_{l1}+\Delta x) \cdot \frac{z_a+d_2}{z_a}, \\
\end{split}
\label{eq:rotation_9}
\end{equation}
where $x_{l1}, x_{l2}, x_{m1}, x_{m2}, x_{r1}, x_{r2}$ are corresponding coordinates of $N_l(t+\Delta t), N_l(t), N_m(t+\Delta t), N_m(t), N_r(t+\Delta t), N_r(t)$ in the dimension of $\textbf{x}$ in recording image space. In facial regions, we can easily find corresponding points $N_l(t), N_m(t)$, which satisfy that $d_1 \ll z_a$ (i.e., $d_1 = 0$) and $x_{m1} > x_{l1}$. In this pattern, Eq.~\ref{eq:rotation_8} can be established. To establish Eq.~\ref{eq:rotation_9}, we only need to find point $N_r(t)$, which satisfies that $x_{r1} < x_{l1}$. According to the derivation above, we can see that there exists cases that $d'_1/d'_2 < d_1/d_2$. And there are also many cases that satisfy $d'_1/d'_2 > d_1/d_2$, which we do not elaborate here. When faces move, the absolute coordinates $x_{l1}, x_{l2}, x_{m1}, x_{m2}, x_{r1}, x_{r2}$ vary, as well as $\beta_1, \beta_2$, leading to the variation of estimated relative depth of three facial points at different moments, which will not occur in the \emph{real} scene. That's to say, if the realistic image plane and recording image plane are not parallel, we can seek cases to detect abnormal relative depth with the help of abnormal facial motion.

\vspace{-10pt}
\subsubsection{Discussion}
One of basis of the elaboration above is that the structure of face is similar to that of the hill, which is complex, dense and undulate. This is interesting and worth being exploited in face anti-spoofing.

Even though we only use some special examples to demonstrate our viewpoints and assumption, they can still prove the reasonability of utilizing facial motion to estimate the relative facial depth in face anti-spoofing task. In this way, the learned model can seek the abnormal relative depth and motion in the facial regions. And our extensive experiment demonstrates our assumption and indicates that temporal depth method indeed improves the performance of face anti-spoofing.

\section{Extra Experiments}
\begin{figure*}[!htbp]
\centering
\subfigure{
\includegraphics[width=0.32\textwidth]{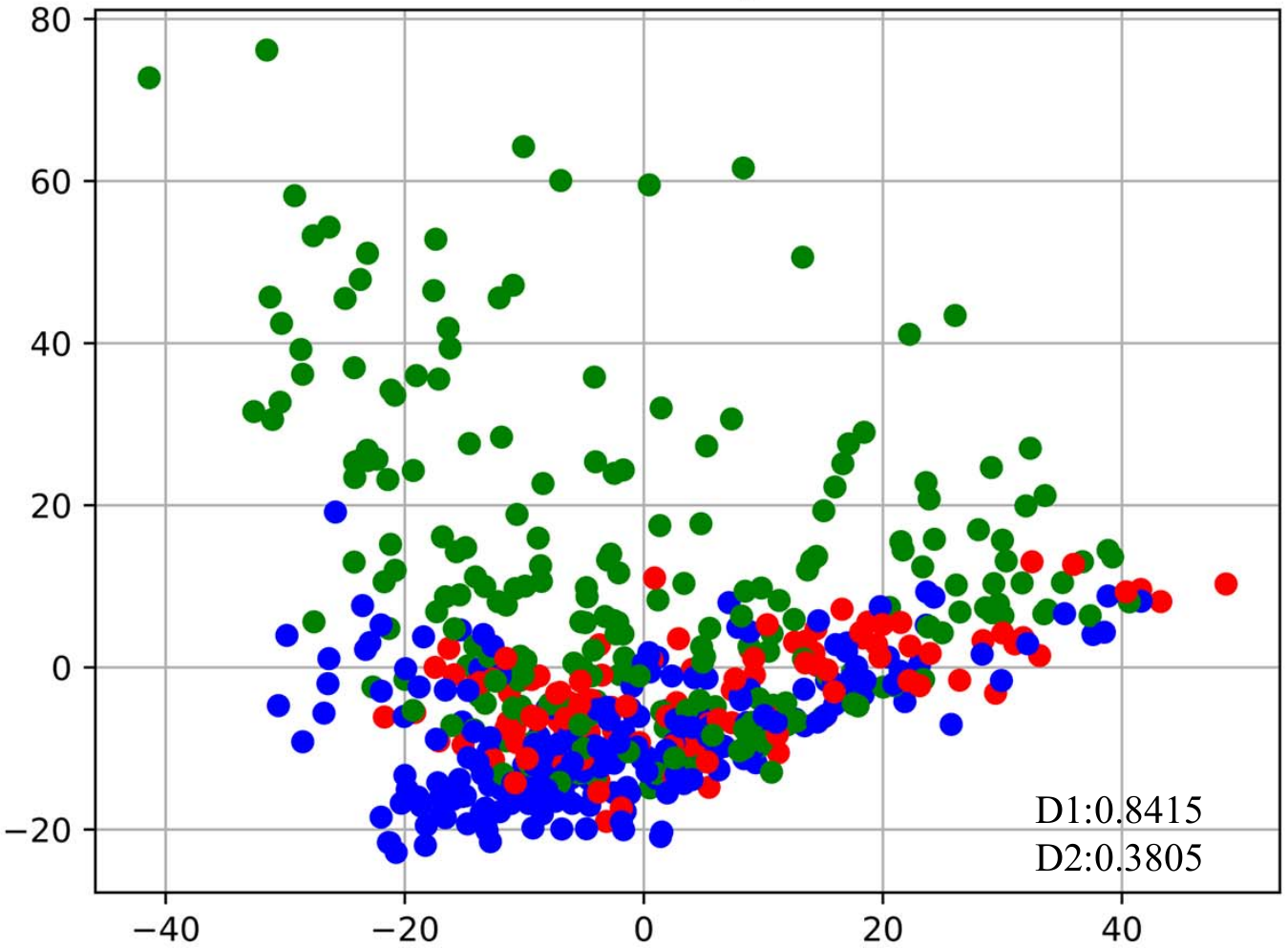}
\label{fig:feature_map_a}
}
\subfigure{
\includegraphics[width=0.32\textwidth]{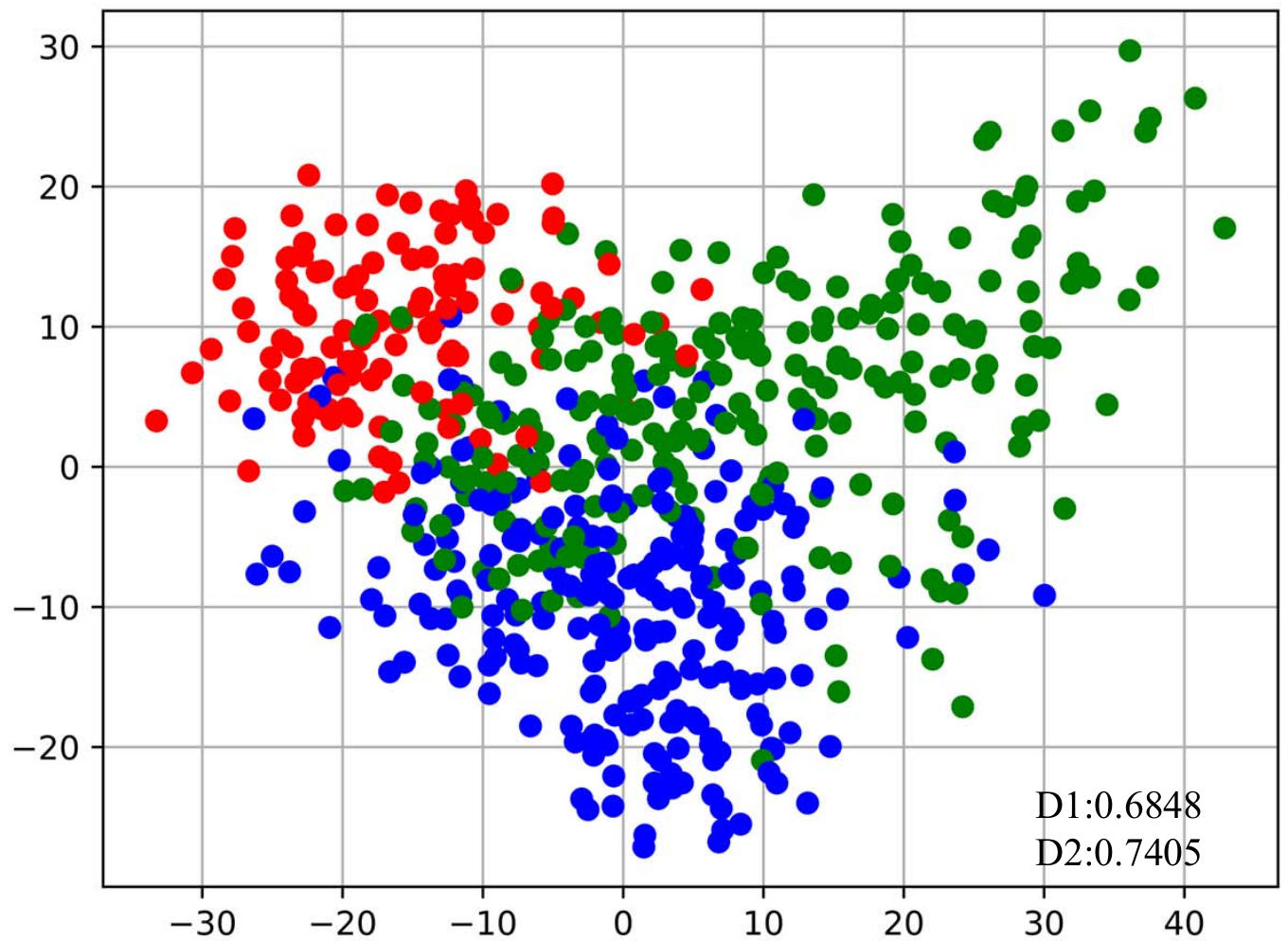}
\label{fig:feature_map_b}
}
\subfigure{
\includegraphics[width=0.32\textwidth]{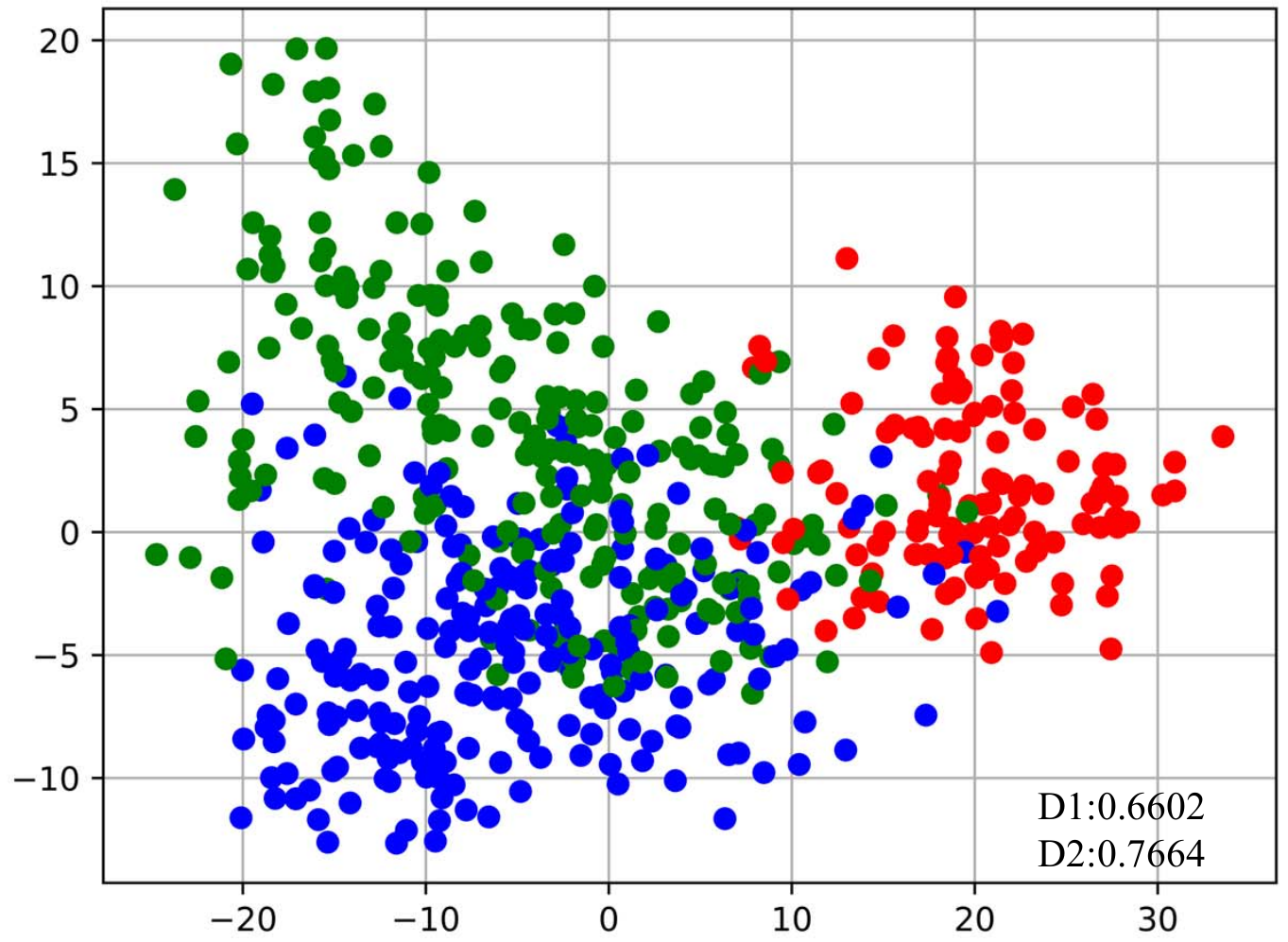}
\label{fig:feature_map_c}
}
\label{fig:feature_map}
\caption{Distribution of features in different modules in our method. We split all samples into three classes. Red, blue, and green points denote samples of real, print attack, and replay attack, respectively.
%D1 is the average distance of samples in the same classes, and D2 is the average distance of samples in different classes.
D1 is intra class distance where smaller value indicates better performance, and D2 is inter class distance where larger value indicates better performance. All distances are calculated after normalizing the features.
}
\vspace{-10pt}
\end{figure*}

\begin{table}
\resizebox{0.48\textwidth}{!}{
\begin{tabular}{|c|c|c|c|c|}%一个c表示有一列\emph{}，格式为居中显示(center)
\hline
Prot. & Method & APCER(\%) & BPCER(\%) & ACER(\%) \\
\hline
\multirow{2}{*}{1} &FAS-BAS~\cite{Liu2018Learning} &-- &-- &10.0 \\
       \cline{2-5} &\textbf{OURs} &2.9 &12.5 &\textbf{7.7} \\
\hline
\multirow{2}{*}{2} &FAS-BAS~\cite{Liu2018Learning} &-- &-- &14.1 \\
       \cline{2-5} &\textbf{OURs} &13.6 &13.3 &\textbf{13.5} \\
\hline
\multirow{2}{*}{3} &FAS-BAS~\cite{Liu2018Learning} &-- &-- &\textbf{13.8$\pm$5.7} \\
       \cline{2-5} &\textbf{OURs} &18.1$\pm$8.7 &10.0$\pm$6.3 &{14.0$\pm$4.5} \\
\hline
\multirow{2}{*}{4} &FAS-BAS~\cite{Liu2018Learning} &-- &-- &10.0$\pm$8.8 \\
       \cline{2-5} &\textbf{OURs} &2.5$\pm$3.8 &12.5$\pm$11.1 &\textbf{7.5$\pm$5.5} \\
\hline
\end{tabular}
}%\resizebox{\textwidth}{!}{
\caption{The results of cross testing from SiW~\cite{Liu2018Learning} to OULU-NPU~\cite{Boulkenafet2017OULU}. } %FAS-BAS is the model in ~\cite{Liu2018Learning}}
\label{tab:SiWOulu}
\vspace{-10pt}
\end{table}

\vspace{-1pt}
\subsection{Comparison}
In Tab.~\ref{tab:SiWOulu}, we show results of cross testing on models trained on SiW~\cite{Liu2018Learning} and tested on OULU-NPU~\cite{Boulkenafet2017OULU}. We can see that our method outperforms FAS-BAS~\cite{Liu2018Learning} on three protocols. Our method even decrease 2.3\%, 0.6\% and 2.4\% of ACER on protocol 1, protocol 2 and protocol 4, respectively. In protocol 3, ACER of our method is 14.0$\pm$4.9\%, which is higher than that of FAS-BAS~\cite{Liu2018Learning} for 0.2\%, which shows that our method is sensitive to the variation of input cameras. This indicates the future research direction.

\subsection{Feature Distribution}
As shown in Fig.~\ref{fig:feature_map}, we present distribution of different kinds of features in our method. The left subfigure is about the feature of single-frame part. The middle subfigure is about the feature of OFF block. The right subfigure is about the feature of ConvGRU. These features are all extracted from our model 1. We can see that the features are getting more discriminative from left to right. The intra class distances are getting nearer from left to right. And the inter class distances are getting further from left to right. This indicates that our features of motion are discriminative in face anti-spoofing.

%\clearpage
%{\small
%\bibliographystyle{ieee}
%\bibliography{refs}
%}

\end{appendix}

\end{document}